%% file: main.tex
\documentclass[10pt,twocolumn,letterpaper]{article}

\usepackage[pagenumbers]{cvpr} % To force page numbers, e.g. for an arXiv version

\input{preamble}

\definecolor{cvprblue}{rgb}{0.21,0.49,0.74}
\usepackage[pagebackref,breaklinks,colorlinks,allcolors=cvprblue]{hyperref}

\usepackage[capitalize]{cleveref}
\crefname{section}{Sec.}{Secs.}
\Crefname{section}{Section}{Sections}
\Crefname{table}{Table}{Tables}
\crefname{table}{Tab.}{Tabs.}

\newcommand{\midsepremove}{\aboverulesep = 0mm \belowrulesep = 0mm}
\midsepremove
\newcommand{\midsepdefault}{\aboverulesep = 0.605mm \belowrulesep = 0.984mm}
\midsepdefault

\def\paperID{11704} % *** Enter the Paper ID here
\def\confName{CVPR}
\def\confYear{2025}

\title{SceneFactor: Factored Latent 3D Diffusion for Controllable 3D~Scene~Generation}

\author{Alexey Bokhovkin\\
Technical University of Munich\\
\and
Quan Meng\\
Technical University of Munich\\
\and
Shubham Tulsiani\\
Carnegie Mellon University\\
\and
Angela Dai\\
Technical University of Munich\\
}

\begin{document}

%%%%%%%%% TEASER
 \twocolumn[{%
 	\renewcommand\twocolumn[1][]{#1}%
 	\maketitle
 	\begin{center}
 		%\vspace{-0.5cm}
 		\includegraphics[width=0.98\linewidth]{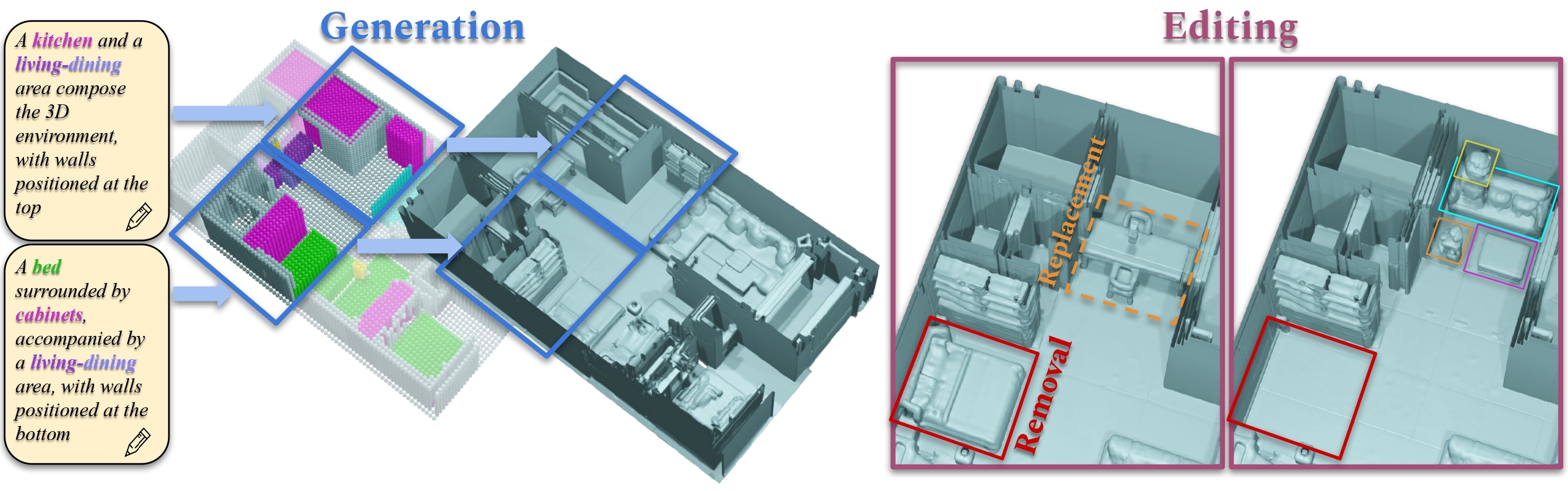}
 		\vspace{-0.4cm}
 		\captionof{figure}{
             SceneFactor factors the complex task of text-guided 3D scene generation into forming a coarse semantic structure, followed by refined geometric synthesis.
             %This enabling high-quality scene generation and intuitive localized editing. 
             Rather than require a learned model to decide the location, type, size, and local geometry of scene elements directly, our generation of a coarse semantic box layout enables training a simpler task of layout-guided geometric synthesis. To achieve this factorized generation, we train semantic and geometric latent diffusion models. Crucially, the proxy semantic map generation enables user-friendly localized editing of generated scenes by editing in the semantic map with simple box operations (by clicking two box corners), without requiring re-synthesis of the full scene.
             Note that input text is colored by semantic categories for visualization purposes only.
 		}
             \vspace{-0.0cm}
 		\label{fig:teaser}
 	\end{center}
 }]

\maketitle
\input{sections/0_abstract}    
\input{sections/1_intro}

\input{sections/2_relatedwork}

\input{sections/3_method}

\input{sections/4_results}
\input{sections/5_conclusion}
\section*{Acknowledgments}
This project was supported by the ERC Starting Grant SpatialSem (101076253), the Bavarian State Ministry of Science and the Arts and coordinated by the Bavarian Research Institute for Digital Transformation (bidt), and the German Research Foundation (DFG) Grant ``Learning How to Interact with Scenes through Part-Based Understanding."
{
    \small
    \bibliographystyle{ieeenat_fullname}
    \bibliography{main}
}

% WARNING: do not forget to delete the supplementary pages from your submission 
\input{sections/X_suppl}

\end{document}

%% file: preamble.tex
\usepackage{url}   
\usepackage{booktabs} % For formal tables
\usepackage{multirow}
\usepackage{comment}
\usepackage{stfloats}
\usepackage{enumitem}
\usepackage{courier}

\usepackage[ruled]{algorithm2e} % For algorithms

\SetAlFnt{\small}
\SetAlCapFnt{\small}
\SetAlCapNameFnt{\small}
\SetAlCapHSkip{0pt}

\usepackage{breqn}

\usepackage[dvipsnames]{xcolor}

\newcommand{\OURS}{SceneFactor}

%% file: sections/0_abstract.tex
\begin{abstract}

%In this paper, we present SceneDiffusion, the first approach for large-scale indoor 3D scene generation and editing from textual prompts.
We present \OURS{}, a diffusion-based approach for large-scale 3D scene generation that enables controllable generation and effortless editing.
\OURS{} enables text-guided 3D scene synthesis through our factored diffusion formulation, leveraging latent semantic and geometric manifolds for generation of arbitrary-sized 3D scenes.
While text input enables easy, controllable generation, text guidance remains imprecise for intuitive, localized editing and manipulation of the generated 3D scenes.
Our factored semantic diffusion generates a proxy semantic space composed of semantic 3D boxes that enables controllable editing of generated scenes by adding, removing, changing the size of the semantic 3D proxy boxes that guides high-fidelity, consistent 3D geometric editing.
Extensive experiments demonstrate that our approach enables high-fidelity 3D scene synthesis with effective controllable editing through our factored diffusion approach.

{\normalfont Project page:} {\normalfont \url{alexeybokhovkin.github.io/scenefactor/}}

\end{abstract}

%% file: sections/1_intro.tex
\section{Introduction}

3D editable generative modeling is crucial to create immersive environments for many applications, such as augmented or virtual reality, video games and films, architectural design, or creating interactive simulations. 
Such content creation is inherently creative by nature, and is typically performed in an iterative process controlled by the user, with the ability to control and edit in localized regions to produce the desired output. 
Thus, a key requirement in generative 3D modeling is an underlying representation that enables such intuitive, localized control and editing for users.

While remarkable advances in 2D generative modeling have been achieved with diffusion-based methods \cite{sohl2015deep,ho2020denoising,rombach2021highresolution,peebles2023scalable} (even enabling controllability through semantic layouts, human poses, or depth \cite{zhang2023adding,mou2023t2i}), 3D generative modeling has largely focused on the unconditional or text-conditioned synthesis of 3D shapes \cite{li2023diffusionsdf,ndf2023shue,erkocc2023hyperdiffusion,cheng2023sdfusion,siddiqui2024meshgpt}, while the more challenging problem of large-scale 3D scene generation remains underexplored. 
Moreover, these methods also tend to lack editability, which is a key requirement of the content creation process -- to be able to edit the generated representation in localized regions without requiring a re-synthesis of the full output. Editable generative approaches often lack the ease of editing operations, requiring the user to specify an accurate editing region boundary~\cite{lugmayr2022repaint, Avrahami_2022_CVPR, sajnani2024geodiffuser} or conduct extensive prompt engineering to avoid editing of undesired regions~\cite{su2022dual, bar2022text2live, deutch2024turbo}.

We thus propose a diffusion-based 3D generative approach for the synthesis of large-scale 3D scenes that enables intuitive, localized editing of the generated 3D representation in two clicks (defining a bounding box) per edited object. 
Key to our approach is a learned, latent semantic feature space which enables localized editability and control of the 3D scene generation.
We learn to map text descriptions of scene regions to 3D semantic layout maps, which then guide the high-fidelity geometric synthesis of scene geometry corresponding to the proxy semantics.
We formulate a two-stage latent semantic diffusion approach, first learning latent semantic and geometric feature spaces through VQ-VAE training. 
The latent semantic space is then modeled by diffusion, condition on text inputs, to produce a proxy semantic map. 
We then model the 3D scene geometry in its latent geometric space, conditioned on the semantic layout maps through spatial cross-attention to enable effective localized modeling of the semantic structure corresponding to geometric outputs.

Edits can then easily be performed in the semantic space by specifying the two points defining a bounding box (which can be automatically filled to match the proxy semantic map characteristics).
To characterize the complexity in 3D scenes and handle larger scales, \OURS{} is trained on scene chunks, which can then be consistently outpainted to generate arbitrary-sized 3D scene outputs.
Experiments show that \OURS{} enables text-guided synthesis as well as intuitive editing in the proxy semantic domain (e.g., adding objects by introducing new semantic boxes, as well as removing, moving, and editing generated objects by manipulating two corners of their semantic boxes).

In summary, our contributions are:
\begin{itemize}
\item the first method for text-guided large-scale 3D scene generation that enables easy, localized spatial editing for generated 3D scenes, performed in several mouse clicks.
\item a latent semantic diffusion approach to enable two-stage generation of semantic and geometric latent manifolds characterizing coarse 3D scene layout and high-fidelity geometric structures, leveraging spatial cross-attention for strong spatial guidance of geometric synthesis.
\item our latent semantic space enables intuitive, localized editing of generated 3D scenes without requiring re-synthesis of the full scene, enabling object addition, removal, replacement, and object manipulation while maintaining global scene consistency.
\end{itemize}

%% file: sections/2_relatedwork.tex
\section{Related Work}

\subsection{3D Shape Generation}

Recent remarkable advances in 2D image generation have re-invigorated research in 3D generative modeling, which has largely focused on shape generation.
Directly inspired by 2D generative models such as latent diffusion models~\cite{rombach2021highresolution}, various methods have been developed to distill information from large, pretrained 2D models for text-to-3D radiance field generation \cite{poole2022dreamfusion, chan2023genvs, zhang2023text2nerf, wang2023prolificdreamer, lin2023magic3d, fantaisa3d}.

Alternatively, many other methods have focused on directly generating 3D shape representations by training on large 3D shape datasets such as ShapeNet~\cite{chang2015shapenet}.
In order to generate significant detail for high-dimensional 3D objects, recent approaches focus on generating compressed latent representations for 3D shapes \cite{autosdf2022, yan2022shapeformer,Wu2016LearningAP,genmodels2020chaudhuri} and efficient mesh representations \cite{siddiqui2024meshgpt,nash2020polygen,khalid2022clipmesh}.
These methods focus on single-object generation in a canonicalized domain, while we focus on large-scale scene generation.

3D diffusion-based methods have also been developed for high-fidelity 3D shape generation.
PVD~\cite{Zhou_2021_ICCV} generates 3D point clouds with a hybrid point-voxel representation.
Diffusion-SDF~\cite{chou2022diffusionsdf}, HyperDiffusion~\cite{erkocc2023hyperdiffusion}, NFD~\cite{ndf2023shue}, and SDFusion~\cite{cheng2023sdfusion} leverage trained 1D, 2D, and 3D representations to more efficiently encode 3D shape geometry. 
While these approaches focus on single object generation, they can be conceptually applied to 3D scene generation by training on crops of 3D scenes.
Our approach not only focuses on a factored diffusion approach for high-fidelity 3D scene generation, but learning a 3D scene representation that enables intuitive, localized editing for content creation scenarios.

% Overview
\begin{figure*}[tp]
\begin{center}
    \includegraphics[width=1.0\textwidth]{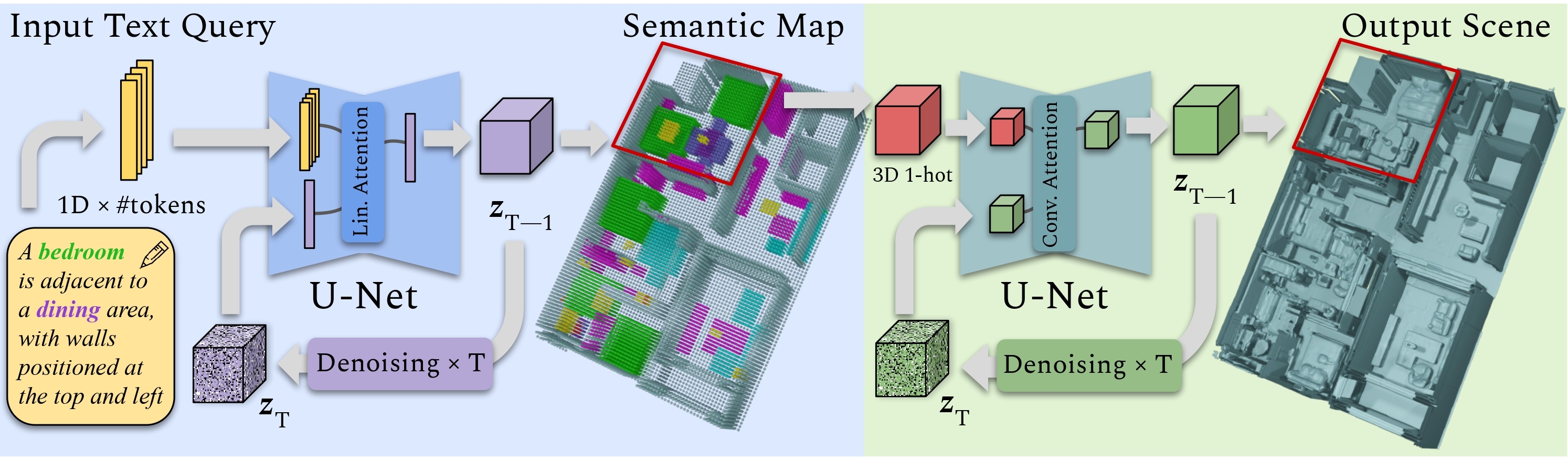}
    \vspace{-0.8cm}
    \caption{
    Method overview. 
    We formulate text-guided 3D scene generation as a factored diffusion process, first generating a coarse semantic box layout representing the text input (left), followed by synthesis of scene geometry corresponding to the generated semantics (right).
    This factorization makes complex 3D scene generation more tractable and enables generation of locally editable 3D scenes, which can be manipulated through box manipulations in the semantic maps.
    Left: Our high-level semantic generation produces a coarse, box-level representation of a scene through latent diffusion on a pretrained semantic manifold, conditioned on text captions. This enables accurate alignment between text input and scene layout, without requiring solving a highly ambiguous generation task for geometric detail. 
    Right: Conditioned on the coarse semantic box map, we use another latent diffusion model to generate 3D scene geometry, enabling spatial semantic grounding of generated scene objects and structures.
    Object categories in the text input are colored for visualization only.
    }
    \vspace{-0.8cm}
    \label{fig:overview}
\end{center}
\end{figure*}

\subsection{3D Scene Generation}

Generating 3D scenes remains significantly more challenging than objects, due to the complexity of scene arrangements, high resolution required to resolve local detail, and strongly varying sizes~\cite{advances2023patil}.
Several approaches have thus relied on the capacity of image generative models to iteratively generate RGB views from text queries in order to form 3D scenes  \cite{hoellein2023text2room, SceneScape, roomdreamer}; this results in impressive local appearance, but the lack of 3D reasoning often results in more incoherent global 3D structures.
GAN-based approaches have enabled 3D-aware scene generation as radiance fields using depth priors~\cite{shi20223daware} or scene layouts~\cite{cc3d2023bahmani}, for improved view synthesis.

A popular approach is to leverage object retrieval in order to create 3D scenes with high-fidelity object structures, and instead synthesize the scene graph of object layouts \cite{tang2024diffuscene, nie2023learning, scenehgn2023gao, li2018grains, Paschalidou2021NEURIPS, wang2018deep, graphvae2023ch, fast3dpriors2022zhang, zhai2023commonscenes, graph2scene2021, AguinaKang2024OpenUniverseIS, wei2023legonet, procthor, yang2023holodeck}. Due to the use of object retrieval, scene geometry remains limited to the object database used for retrieval.
Most similar to our approach are several recent approaches, DiffInDScene~\cite{ju2023diffroom}, BlockFusion~\cite{Wu2024blockfusion}, SemCity~\cite{lee2024semcity} and XCube~\cite{ren2023xcube}, which have been developed directly for large-scale scene generation, leveraging more flexible 3D representations unrestricted by object retrieval.
In particular, BlockFusion produces a scene in a sliding window fashion, conditioned on a given layout, employing a single triplane latent diffusion stage.
XCube generates the structure of an entire scene at once, without relying on sliding windows, instead generating a scene in a hierarchically coarse-to-fine fashion. 
Our approach also takes a chunked approach to scene generation, enabling large-scale synthesis of 3D scenes by chunk-based outpainting.
However, in contrast to state-of-the-art 3D diffusion approaches that focus on direct scene generation that do not enable editing of scene outputs, we develop a factored diffusion approach to enable both high-fidelity geometric synthesis while enabling localized editing of output scenes.

% Retrieval-based or code dict, scene graph \cite{scenehgn2023gao} \cite{li2018grains} \cite{Paschalidou2021NEURIPS} \cite{wang2018deep} \cite{graphvae2023ch} \cite{fast3dpriors2022zhang} \cite{zhai2023commonscenes} \cite{graph2scene2021}
% Image-based \cite{ren2022look} \cite{Wiles2019SynSinEV}  \cite{Rockwell2021} \cite{Xu_2021_CVPR}
% 3D-aware GAN \cite{shi20223daware}
% layout gan \cite{cc3d2023bahmani} \cite{Vidanapathirana2021Plan2Scene}
% transformer vq \cite{autosdf2022} \cite{yan2022shapeformer}
% clip nerf \cite{Lee2022UnderstandingPC} 
% clip mesh \cite{khalid2022clipmesh}
% nerf \cite{Schwarz2020NEURIPS} \cite{Niemeyer2020GIRAFFE}

% retrieval scene graph \cite{tang2024diffuscene}
% Nerf \cite{poole2022dreamfusion} \cite{chan2023genvs} \cite{zhang2023text2nerf} \cite{wang2023prolificdreamer} \cite{lin2023magic3d} \cite{fantaisa3d}
% View uplifting \cite{roomdreamer} \cite{SceneScape} \cite{hoellein2023text2room}
% llm retrieval shape program \cite{AguinaKang2024OpenUniverseIS}
% geometry diffusion \cite{cheng2023sdfusion} \cite{chou2022diffusionsdf} \cite{Zhou_2021_ICCV} \cite{li2023diffusionsdf} \cite{ju2023diffroom} \cite{Wu2024blockfusion} \cite{ndf2023shue} \cite{ren2023xcube}

\subsection{3D Object and Scene Editing}

Generating controllable 3D object or scene representations has largely focused on conditional generative modeling formulations, using input text, images, or partial scans to guide output synthesis.
For 3D shapes, methods such as AutoSDF~\cite{autosdf2022} and ShapeFormer~\cite{yan2022shapeformer} enable 3D shape generation conditioned on image or partial 3D object inputs.
3D diffusion models can also be formulated as conditional diffusion models to enable text- or image-based 3D generation \cite{cheng2023sdfusion,li2023diffusionsdf,jun2023shape,zhao2023michelangelo}.
Research on 3D scenes has also emphasized conditional generation, largely based on text and/or scene layout information to generate 3D scenes \cite{fang2023ctrl,Wu2024blockfusion,schult24controlroom3d,Yan2024FrankensteinGS}.
While such conditional generative approaches enable high-level control over generated outputs based on adapting the input text, image, or layout, they would require re-synthesis of the generated output for adapted inputs, making localized editing challenging.

For 3D shapes, several approaches have been developed to enable more fine-grained localized shape editing, through local attention~\cite{zheng2023lasdiffusion} or part-based reasoning \cite{editvae2022li,nakayama2023difffacto,Tertikas2023CVPR}.
Our approach formulates a factored diffusion approach to enable localized editing of generated 3D scenes.

%controlled image and texture generation \cite{Avrahami_2022_CVPR} \cite{Avrahami_2023_CVPR} \cite{multidiffusion2023bartal} \cite{bashkirova2023masksketch} \cite{brooks2022instructpix2pix} \cite{chen2023text2tex} \cite{hertz2022prompt} \cite{lhhuang2023composer} \cite{mou2023t2i}

%controlled 3D generation \cite{fang2023ctrl} \cite{jun2023shape} \cite{zhao2023michelangelo} \cite{Liu2023MeshDiffusion} \cite{zheng2023lasdiffusion} \cite{editvae2022li} \cite{pointto3d2023yu} \cite{nakayama2023difffacto} \cite{Tertikas2023CVPR}

%geometry diffusion \cite{cheng2023sdfusion} \cite{Zhou_2021_ICCV} \cite{li2023diffusionsdf} \cite{Wu2024blockfusion}

%indoor scene from layout \cite{schult24controlroom3d}
%indoor scene \cite{Yan2024FrankensteinGS}

%% file: sections/3_method.tex
\section{Method}

\OURS{} is a factored diffusion-based approach that generates large-scale 3D indoor scenes from text, using a proxy 3D semantic space to enable synthesis of high-fidelity, controllable 3D scenes. 
From an input text caption $\tau$, we first synthesize a coarse 3D semantic layout $S$, representing a scene as 3D semantic boxes corresponding to the text $\tau$.
Based on semantic layout $S$, we then synthesize output scene geometry $G$.
This factors the complex 3D scene generation process to high-level structural generation, followed by synthesis of geometric detail, enabling high-fidelity synthesis.
Moreover, this enables the output scene $G$ to be locally edited by simple manipulations performed on $S$.
Both factored semantic and geometric representations are generated through conditional latent diffusion to produce compressed feature representations $S$ and $G$.

In order to synthesize large-scale scene environments, training is performed on scene chunks, and a 3D scene is generated chunk-by-chunk through outpainting.
A set of input text descriptions  $\{\tau_k\}_{k=1}^{N_c}$ provides high-level user control over the scene chunk generation, where $N_c$ is the number of chunks to generate for an output scene.

We first describe the chunking of scenes for training our factored latent spaces as well as diffusion training in Sec.~\ref{subsec:chunks}.
We then optimize our factored latent semantic and geometric spaces (Sec.~\ref{subsec:latenttraining}), followed by diffusion training over these spaces (Sec.~\ref{subsec:diffusion}).
Finally, 3D scenes are synthesized by chunk-by-chunk outpainting (Sec.~\ref{subsec:outpainting}), and our factored generation approach enables localized 3D scene editing of synthesized scenes (Sec.~\ref{subsec:editing}).

\begin{figure}
\begin{center}
    \includegraphics[width=0.99\linewidth]{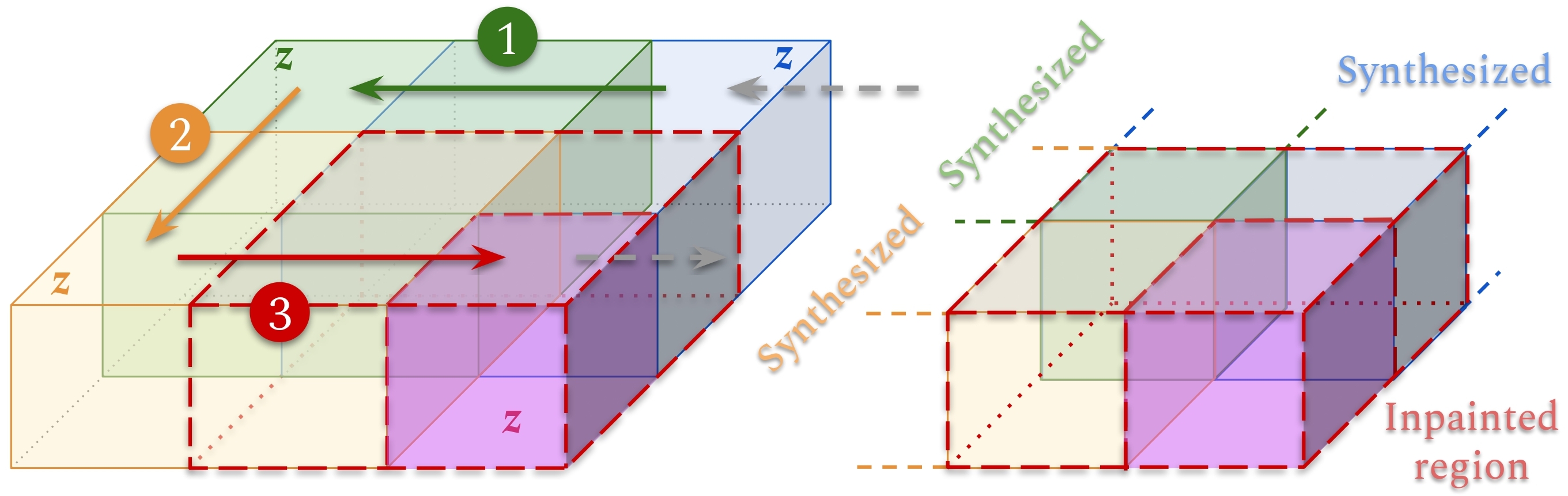}
    \vspace{-0.4cm}
    \caption{Chunk-based 3D scene generation. Left: Chunks for a scene are generated in sliding-window fashion (1-2-3), with overlap between generated chunks to ensure scene consistency along boundaries.
    Right: Synthesis of a chunk (chunk 3) is based on regions of previously generated chunks (1,2).  The purple incomplete region is then synthesized by inpainting based on the previously generated blue, green, and yellow regions.
    }
    \vspace{-0.8cm}
    \label{fig:chunks}
\end{center}
\end{figure}

\subsection{Chunk-based 3D Scene Generation}
\label{subsec:chunks}

To produce 3D scenes of arbitrary sizes, we train our approach on scene chunks and synthesize output scenes in chunk-by-chunk fashion.
As shown in Fig.~\ref{fig:chunks}, a train scene $X$ is chunked into $N_c$ chunks in sliding-window fashion along the $x$ and $y$ axes ($z$ remains a constant height).
Chunks are generated with half-chunk overlap.

For a scene $X$, we then generate $N_c$ chunks with text captions $\{\tau_k\}_{k=1}^{N_c}$, and corresponding semantic grids $\{S_k\}_{k=1}^{N_c}$ and geometric grids $\{G_k\}_{k=1}^{N_c}$.
Each semantic chunk $S_k$ contains a grid of one-hot-encodings of semantic boxes for each class category, where the first channel corresponds to free space, the second to wall/floor and the remaining 8 channels for object categories. The object categories are shown in Fig.~\ref{fig:editing}. Each geometric chunk $G_k$ describes a truncated unsigned distance field representation of the scene geometry in the chunk. We use cubic-sized chunks for the VQ-VAE training, with chunks twice as large (except in the up direction) for diffusion training.

To generate 3D scenes of arbitrary sizes, we describe chunk-by-chunk synthesis in Sec.~\ref{subsec:outpainting}, first generating the full semantic scene map based on a set of text captions, and then generating refined scene geometry, leveraging our factored generation process to disentangle the complex task of 3D scene generation into high-level semantic mapping followed by finer-grained geometric synthesis.

\subsection{Factored Semantic and Geometric Latent Optimization}
\label{subsec:latenttraining}

\OURS{} leverages dual semantic and geometric latent spaces for factored scene generation, enabling high-fidelity and editable 3D scene synthesis through disentangling the 3D scene generation task.
To obtain both latent semantic and geometric spaces, we first optimize two models to encode compressed feature representations $f_S$ and $f_G$ that can be decoded to semantic and geometric chunks $S$ and $G$.

Geometric distance field chunks $G\in \mathbb{R}^{128 \times 64 \times 128}$ are spatially compressed by a factor of 4 to $f_G\in\mathbb{R}^{32 \times 16 \times 32}$. Here, the latent space size is designed to be as small as possible to encourage effective generative modeling while still being able to decode to high-fidelity geometry. Semantic one-hot chunks $S\in\mathbb{Z}^{c \times 32 \times 16 \times 32}$, where $c=10$ denotes the number of class categories, are also spatially compressed by a factor of 4 to $f_S\in\mathbb{R}^{8 \times 4 \times 8}$.

%As already mentioned, we leverage chunked fragmentation to enable the generation of scenes of various sizes and aspect ratios and even spatially unlimited scenes. Along with this advantage, this partitioning scheme allows us to train the model on chunked data, which provides a combinatorically unlimited number of chunks that can be extracted from the training scenes. Furthermore, it becomes possible to regulate the chunk size to efficiently balance between a chunk context window and computational speed during training and testing.

%The scene chunks $\{s_i\}$ are first transformed into uDF representation $\{u_i\} \in \mathbb{R}^{128 \times 64 \times 128}$ with the resolution of $?$ cm using~\cite{} and truncated using the distance of $?$ cm. Though these chunks can be reconstructed into high-quality meshes, we found that training diffusion models on samples with a high number of parameters is significantly difficult. Training the diffusion models on this data leads to a highly limited diversity of semantic and geometric chunks and lower quality. Corresponding chunks of semantic maps are one-hot encoded as $\{w_i\} \in \mathbb{Z}^{c \times 32 \times 16 \times 32}$ and have the resolution of $?$ cm, where $c=10$ is the number of semantic classes including scene layout and free space.

In order to construct latent spaces that are memory efficient and produce smooth manifolds for efficient generation and editing using diffusion, we optimize for both semantic and geometric latent spaces using 3D VQ-VAEs~\cite{van2017neural}.
Empirically, we found that using VQ-VAEs enabled high spatial compression with low feature dimensionality, enabling significant parameter reduction in encoding high-dimensional 3D data.
Note that both latent semantic and geometric feature grids $f_S$ and $f_G$ maintain compressed feature representations with feature dimensionality of 1, enabled through VQ-VAE latent space training.
In particular, we maintain 3D latent grids for both semantics and geometry to enable learning spatial correlation with the decoded spatial 3D domains.
This enables our localized semantic editing of generated 3D scenes.
%Therewith, 3D latent grids ensure a correlation between semantic and geometric latent representation and 3D physical space, which enables accurate attention to semantic conditions and localized changes during the denoising process of diffusion models. 

%We train the compression of geometric $\{u_i\}$ and semantic chunks $\{w_i\}$ using the VQ-VAE model with 3D convolutional layers in the encoder and decoder, where geometric chunks are encoded into the latent grids $\{u^{lat}_i\} \in \mathbb{R}^{32 \times 16 \times 32}$ and semantic chunks are encoded into the grids $\{w^{lat}_i\} \in \mathbb{R}^{8 \times 4 \times 8}$. Both geometric and semantic latent grids have only 1 feature channel. Given the $\mathcal{E}^{geo}$ and $\mathcal{D}^{geo}$ as the encoder and the decoder of the geometric VQ-VAE model, the loss for the geometric reconstruction is shown as follows:
We then train a fully-convolutional 3D VQ-VAE for our geometric latent space, with encoder $\mathcal{E}^{G}$ and decoder $\mathcal{D}^{G}$ optimized for geometric reconstruction:
\begin{equation}
\mathcal{L}^{\textrm{geo}} = \|G - \mathcal{D}^{G}(\mathcal{E}^{G}(G))\|_1 + \mathcal{L}^{\textrm{quant}}(f_G),
\end{equation}
where $\mathcal{L}^{\textrm{quant}}$ is the standard VQ-VAE quantization loss. 
Analogously, given the semantic encoder $\mathcal{E}^{S}$ and decoder $\mathcal{D}^{S}$ for the semantic 3D VQ-VAE, the semantic space is trained using the loss:
\begin{equation}
\mathcal{L}^{\textrm{sem}} = \mathcal{L}^{NLL}(S, \mathcal{D}^{S}(\mathcal{E}^{S}(S))) + \mathcal{L}^{\textrm{quant}}(f_S),
\end{equation}
\begin{dmath}
\mathcal{L}^{NLL}(S, \mathcal{D}^{S}(\mathcal{E}^{S}(S))) = -\sum_{k=1}^c [S]_k \log [\text{softmax}(\mathcal{D}^{S}(\mathcal{E}^{S}(S)))]_k,
\end{dmath}
where $[\cdot]_k$ denotes the $k^{\textrm{th}}$ feature channel corresponding to class $k$, and $\mathcal{L}^{\textrm{quant}}$ is the standard VQ-VAE quantization loss.
The 3D latent encodings of semantics and geometry enables improved reconstruction as well as enabling localized editing based on manipulation of the semantic maps.
%Compared to the previous works, Diffusion-SDF~\cite{?}, BlockFusion~\cite{?} and NFD~\cite{?}, which encode the geometry into one-dimensional or triplanes latent representations, 3D latent grids achieve greater reconstruction accuracy with faster convergence speed and at the same time enable the ease of editing, which will be explained in the next sections.

\subsection{Factored 3D Scene Diffusion}
\label{subsec:diffusion}

Having obtained our factored latent semantic and geometric spaces, we can then train diffusion models, first to generate coarse semantic maps, and then to produce refined geometric synthesis.
%Since we have trained the geometric and semantic VQ-VAE models, the smooth latent grid manifolds with efficient parameterization can be accessed by Diffusion models. Inspired by Stable Diffusion~\cite{?}, the compression of the geometric and semantic chunks removes high-frequency details but learns semantic variation, which provides significantly greater semantic consistency and diversity for the likelihood-based generative models. Before going into detail about our two-stage diffusion approach we first introduce the brief preamble on Diffusion Probabilistic models.
We adopt denoising diffusion probabilistic modeling (DDPM~\cite{Ho2020DenoisingDP}) to denoise the semantic and geometric feature representations $f_S$ and $f_G$ from isotropic Gaussian noise in an iterative process. More specifically, DDPM takes a sample $x_0$ from the input data distribution $q(\mathbf{x})$ and iteratively adds small portions of Gaussian noise to obtain a sequence $x_1, x_2, \dots, x_T$ until $x_T$ reaches approximately an isotropic Gaussian $\mathcal{N}(\mathbf{0}, \mathbf{I})$. According to DDPM~\cite{Ho2020DenoisingDP}, the element $x_t$ of this Markov Chain can be produced using the forward step:
\begin{equation}
q(x_t|x_{t-1}) \sim \mathcal{N}(x_t ; \sqrt{1 - \beta_t} x_{t-1}, \beta_t \mathbf{I}),
\end{equation}
where $\beta_t$ is a variance schedule. During training, DDPM reverses the diffusion process and learns to predict the denoised sample $x_0$ from noisy $x_t$ using a model $p_{\theta}$, often represented as a neural network.

With $\alpha_t := 1 - \beta_t$, $\overline{\alpha}_t := \prod_{s=0}^t\alpha_s$ and $\epsilon \sim \mathcal{N}(\mathbf{0}, \mathbf{I})$, we can sample $x_t$ directly from $x_0$: 
\begin{equation}
x_t = \sqrt{\overline{\alpha}_t}x_0 + \sqrt{1 - \overline{\alpha}_t}\epsilon.
\end{equation}
%
% We apply Bayes’ rule to rearrange the terms and represent $\Tilde{\mu}(x_t, x_0) = \frac{\sqrt{\alpha}_t (1 - \overline{\alpha}_{t-1})}{1 - \overline{\alpha}_t}x_t + \frac{\sqrt{\overline{\alpha}_{t-1}}\beta_t}{1 - \overline{\alpha}_t}x_0$. The closed-form parameterization of $x_t$ yields $\Tilde{\mu}_t = \frac{1}{\sqrt{\alpha_t}}(x_t - \frac{1 - \alpha_t}{\sqrt{1 - \overline{\alpha}_t}}\epsilon_t)$. According to these derivations, we can train the neural network $p_{\theta}$ to learn $\Tilde{\mu}_t$, or alternatively, $\epsilon_t$ by rearranging the terms. 
%
To recover the signal instead of the added noise, we follow~\cite{cheng2023sdfusion, chou2022diffusionsdf} for the reverse process. In our implementation, we use the $v_t$ parameterization as $v_t = \sqrt{\overline{\alpha}_t}\epsilon_t - \sqrt{1 - \overline{\alpha}_t}x_t$.

For our text-to-semantic diffusion model, we construct a 3D variant of OpenAI LDM~\cite{openaidiffusion2021dhariwal} for the main model, and use a transformer with the BERT~\cite{Devlin2019BERTPO} tokenizer to encode an input text query into a tokenized sequence of features. To condition the diffusion model, we apply attention where text features $\tau_i$ are treated as values and latent grids as queries and keys. Thus, the objective for the latent semantic diffusion model $\Psi_{S}$ is the following:
\begin{equation}
\mathcal{L}_{LDM, \textrm{sem}} = \|\Psi_{S}(f_{S,t}, t, \tau_i) - v_{S,t}\|_1,
\end{equation}
where $t$ denotes the timestep of the diffusion process, $f_{S,t}$ is the noisy version of the feature $f_S = f_{S,0}$ and $v_{S,t} = \sqrt{\overline{\alpha}_t}\epsilon_t - \sqrt{1 - \overline{\alpha}_t}f_{S,t}$ is the $v$-parameterization.

The semantic-to-geometry diffusion model is trained analogously with semantic maps as a condition. However, we have found that increasing awareness from local neighbourhoods is essential to capture correlations between semantic map condition and latent grid efficiently. To this aim, we modify linear layers that predict queries, keys and values as features of every separate geometric latent grid or semantic map cell. These linear layers can be viewed as convolutions with a window size of 1 and thus we employ convolutional-based attention modules with a window size of 3, where semantic maps $S$ serve as values and latent grids as queries and keys. The second-stage diffusion model $\Psi_G$ objective is analogous to the first-stage model:
\begin{equation}
\mathcal{L}_{LDM, geo} = \|\Psi_G(f_{G, t}, t, f_S) - v_{G,t}\|_2,
\end{equation}
where $f_{G,t}$ is the noisy version of the feature $f_G = f_{G,0}$ and $v_{G,t} = \sqrt{\overline{\alpha}_t}\epsilon_t - \sqrt{1 - \overline{\alpha}_t}f_{G,t}$ is the $v$-parameterization.

% Editing
\begin{figure*}[tp]
\begin{center}
    \includegraphics[width=0.97\textwidth]{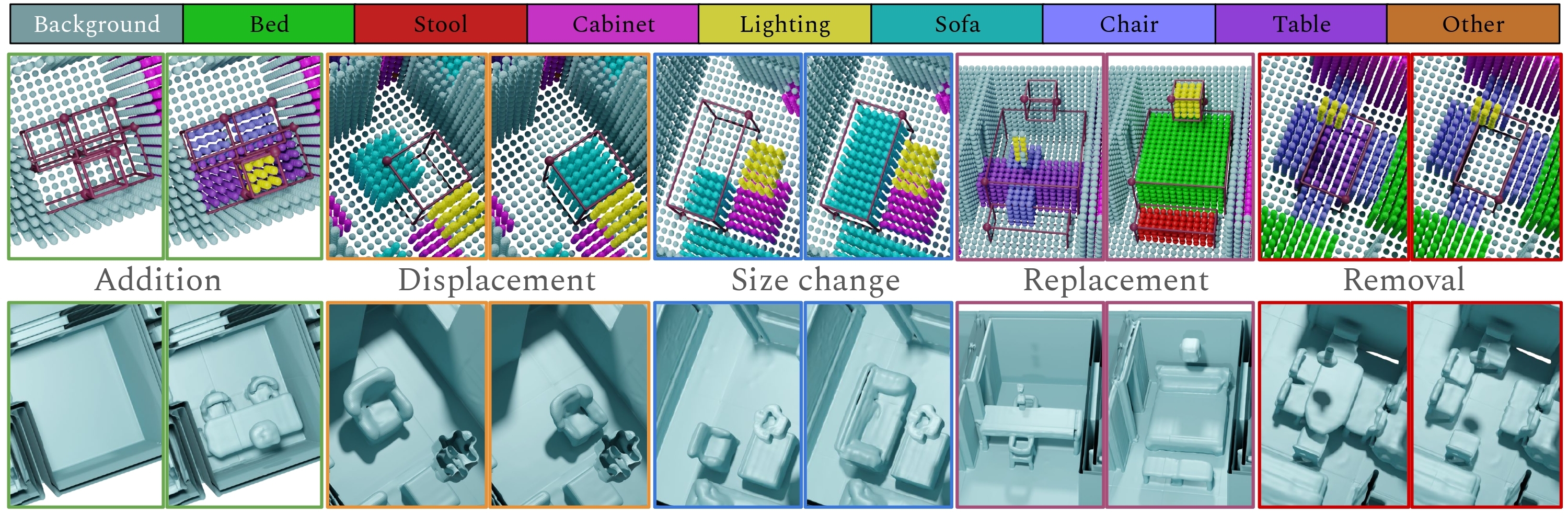}
    \vspace{-0.3cm}
    \caption{Scene editing. \OURS{} enables seamless localized editing through easy manipulation of the 3D semantic box map. 
    We demonstrate the addition of objects (adding boxes), moving objects (moving an existing semantic box), changing object size (scaling an existing semantic box), replacing objects (replacing an existing object box with a new one of a different category), and removing objects (removing an existing semantic box).
    Note that the rest of the 3D scene remains consistent outside of the editing region.
    }
    \label{fig:editing}
    \vspace{-0.75cm}
\end{center}
\end{figure*}

\subsection{Outpainting Large-scale 3D Scenes}
\label{subsec:outpainting}

We train our factored diffusion models on fixed-size scene chunks; however, 3D scenes can have arbitrary spatial sizes.
Thus, we must expand a generated chunk to form a full 3D scene. % generated portion of a scene to its complete form. 
We generate such 3D scenes in a chunk-based sliding-window fashion, using overlaps between neighboring windows.
From one or several already predicted chunks, we formulate the next chunk generation using its corresponding chunk text condition for inpainting, similar to  RePaint~\cite{lugmayr2022repaint}.
We first outpaint semantic chunks, and then refine them to synthesize the corresponding scene geometry.

In Fig.~\ref{fig:chunks}, we show the step-by-step generation process for a simple example scene. 
The first (blue) chunk of latents is generated conditioned only on its text description.
Our sliding window then moves along the direction of the arrows.
The green chunk is then synthesized by inpainting, using the overlapping half of the already synthesized blue region (inpainting only the missing half).
Inpainting is performed by modifying the denoising step, where instead of the classical denoising step formulation $f_{S,t-1} \sim \mathcal{N}(\Tilde{\mu}_{\theta}(f_{S,t};t),\Sigma_{\theta}(f_{S,t};t))$ for step $t$, the inpainting modification is applied:
\begin{equation}
f_{S,t-1}^{\text{known}} \sim \mathcal{N}(\sqrt{\overline{\alpha}_t}f_S^{\text{known}}, (1-\overline{\alpha}_t)\mathbf{I}),
\end{equation}
\begin{equation}
f_{S,t-1}^{\text{unknown}} \sim \mathcal{N}(\Tilde{\mu}_{\theta}(f_{S,t};t),\Sigma_{\theta}(f_{S,t};t)),
\end{equation}
\begin{equation}
f_{S,t-1} = m \odot f_{S,t-1}^{\text{known}} + (1 - m) \odot f_{S,t-1}^{\text{unknown}},
\end{equation}
where $f_S^{\text{known}}$ is a previously generated part of the scene, and $m$ is a binary mask aligned with the currently generated chunk with ones denoting the known region within a chunk. 
Similarly, the yellow chunk is then inpainted given the already synthesized green half.
The next chunk to be synthesized is highlighted in red, which shares overlap with the already synthesized blue, green, and yellow chunks.
The missing purple region of the red chunk is then inpainted.

Since we use sliding windows with a step size of half of the horizontal chunk size, the unknown region to be inpainted is always either 25\%, 50\%, or 100\% of the full chunk size in terms of number of parameters. 
Once the semantic map latent representation of a scene is fully outpainted, we traverse it using the same path of chunks and decode every chunk latent grid into a semantic map chunk with the VQVAE decoder $\mathcal{D}^S$. During this decoding process, the next semantic chunk always overwrites the regions that were previously decoded. The full scene geometric latent representation is outpainted analogously using the generated semantic maps as a condition. However, to obtain the full geometric scene representation we do not perform chunkwise decoding but decode the entire scene geometric latent grid using $\mathcal{D}^G$ to avoid seams between chunks.

\subsection{Localized 3D Scene Editing}
\label{subsec:editing}

Crucially, our factored diffusion approach, disentangling 3D scene generation into coarse semantic synthesis followed by geometric refinement, enables various localized scene edits that can be performed by easy semantic box manipulation of the proxy semantic map representation in just a few mouse clicks.
We demonstrate five example scene edits (object addition, removal, replacement, size changing, and displacement) in Fig.~\ref{fig:editing}.
We also choose equal resolutions of geometric chunk latents $f_G$ and semantic chunk conditions $S_k$, ensuring their exact spatial alignment. % and that one region within $S_k$ corresponds to the same region in $f_G$. 
This enables edits to propagate seamlessly from $S_k$ to $f_G$, improving scene consistency after editing.

Edits are performed as follows, as simple, user-friendly box manipulations of the coarse semantic representation, specifying the two opposite box corners and possibly a new semantic class. For semantic grid $S$ and corresponding geometric latent grid $F_G$:
\begin{itemize}[leftmargin=*,topsep=0pt]
    \item {\bf Object addition}: is performed by adding semantic bounding boxes into an empty editing region $\mathcal{R}_S$ of the scene semantic map $S$. 
    %We initialize the scene geometric latent grid $F_G$ (comprised of geometric chunk latents $f_G$) with geometric features corresponding to the scene prior to the performed editing. 
    We fill only $\mathcal{R}_S$ in the grid $F_G$ with Gaussian noise and re-generate geometry for it. 
    \item {\bf Object removal}: is performed by locating a 3D grid region $\mathcal{R}_S$ corresponding to the object to be removed, and deleting all semantic voxels that belong to it. The same region of the grid $F_G$ is assigned to Gaussian noise and re-synthesized.
    \item {\bf Object replacement}: is performed by replacing the 3D grid region $\mathcal{R}_S$ with a semantic box corresponding to the category of object desired as a replacement. The same region of the grid $F_G$ is then assigned to Gaussian noise and re-synthesized.
    \item {\bf Changing object size}: we first select an object in semantic map $S$ and either increase (by adding new voxels to a box of the same category) or decrease (removing voxels by axis-aligned slices) its box size. 
    We consider the editing region $\mathcal{R}_S$ to be the union of the original box and new box, and similarly re-assign the corresponding region of $F_G$ to Gaussian noise for re-synthesis. Note that since we operate on a semantic rather than instance layout, size increases much larger than the likely size of an object tend to produce multiple objects to fill the size increase.
    \item {\bf Moving an object}: an object is selected by selecting a box region $\mathcal{R}^1_S$ in the semantic map $S$; this box is then translated to the new region $\mathcal{R}^2_S$ of the same size as $\mathcal{R}^1_S$. The geometric features are analogously translated from $\mathcal{R}^1_S$ to $\mathcal{R}^1_S$ of the geometrical feature grid $F_G$. The initial region $\mathcal{R}^1_S$ of $F_G$ is then filled with Gaussian noise, and both regions are re-synthesized.
\end{itemize}
%All the regions ($\mathcal{R}_S$ or $\mathcal{R}^1_S$), where the Gaussian noise was assigned, are regenerated using the diffusion model $\Psi_G$.

%% file: sections/4_results.tex
\section{Experimental Results}

\subsection{Experimental Setup}

\textbf{Datasets.} We train and evaluate our method using a combination of the 3D-FRONT~\cite{fu20213dfront} and 3D-FUTURE~\cite{fu20213dfuture} datasets. 3D-FUTURE contains $>$15,000 3D furniture models from 34 class categories. 3D-FRONT has 18,968 3D indoor scenes furnished with 3D-FUTURE objects. We obtain ~3 million 3D crops of sizes 2.7m and 5.4m to train our VQ-VAE and diffusion models (voxel size 4.2cm). 
After filtering out empty or near-empty scenes, we use a train/test split of 6000/250. 

We obtain two types of captions for scene chunks, where the first set of captions is automatically generated from the 3D-FRONT object annotations in a template-based fashion, and the second set is refined from the first using Qwen1.5~\cite{qwen1.5}. For further information about data processing and caption generation, we refer to the supplemental.

\begin{figure*}[tp]
\begin{center}
    \includegraphics[width=0.97\textwidth]{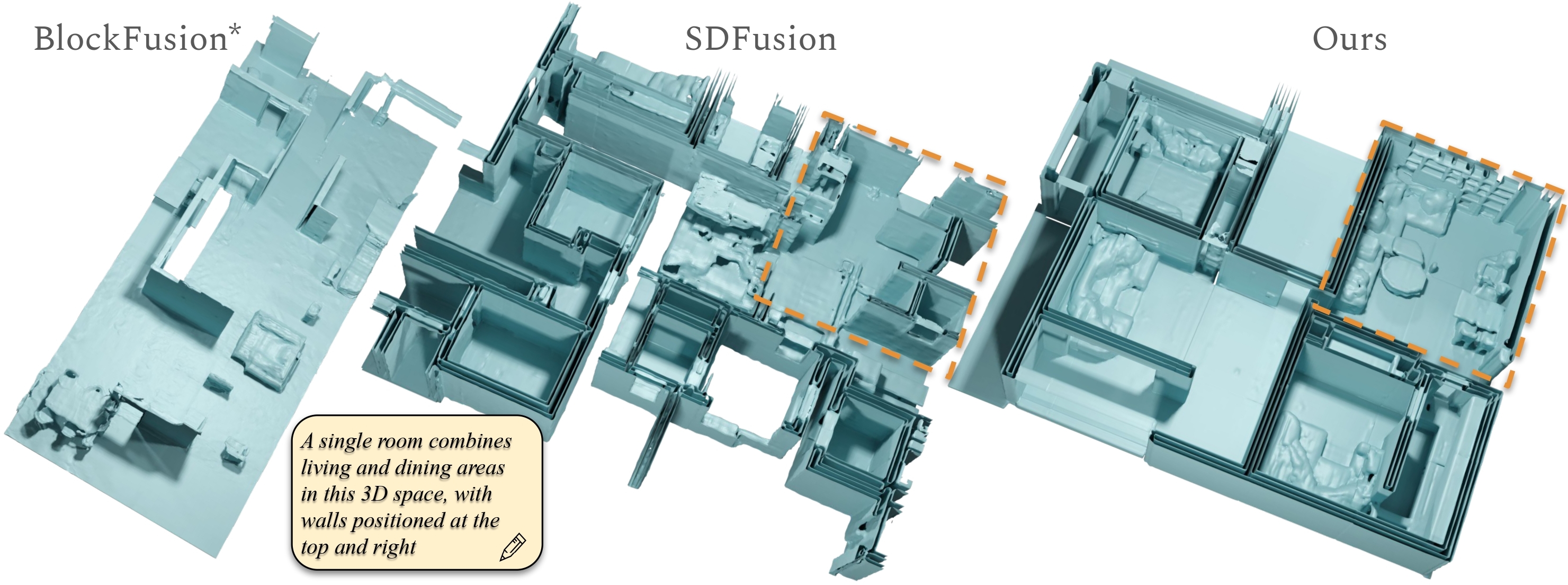}
    \vspace{-0.3cm}
    \caption{Qualitative comparisons to state-of-the-art diffusion-based 3D scene generative approaches BlockFusion~\cite{Wu2024blockfusion}, and SDFusion~\cite{cheng2023sdfusion}. Our approach produces improved scene geometry and more cohesive global scene structure with consistent walls compared to baselines. *Note that results for BlockFusion are generated unconditionally.
    }
    \label{fig:scenes_qual}
    \vspace{-0.7cm}
\end{center}
\end{figure*}

\textbf{Implementation Details.} Our method is trained with an Adam~\cite{KingBa15} optimizer with learning rates 1e-4 and 2e-4 for the semantic and geometric VQ-VAEs. We use AdamW~\cite{Loshchilov2017DecoupledWD} with a learning rate 1e-5 for both semantic and geometric latent diffusion models. The semantic and geometric VQ-VAEs are trained on 2 NVIDIA A6000s each for 320k and 160k iterations ($\sim$ 50 hours) until convergence. The diffusion models are trained on 2 NVIDIA A100s each for 400k iterations ($\sim$ 100 and 150 hours, respectively). 

\begin{table}[bp]
\centering
\vspace{-0.25cm}
\resizebox{0.48\textwidth}{!}{
\begin{tabular}{l|cc|cc|cc|cc|cc|cc}
\toprule
    \multirow{2}{*}{Method} & \multicolumn{6}{c|}{Independent chunks} & \multicolumn{6}{c}{Scene chunks} \\
\cmidrule{2-13}
    & \multicolumn{2}{c}{MMD $\downarrow$} & \multicolumn{2}{c}{COV $\uparrow$} & \multicolumn{2}{c|}{1-NNA (0.5)} & \multicolumn{2}{c}{MMD $\downarrow$} & \multicolumn{2}{c}{COV $\uparrow$} & \multicolumn{2}{c}{1-NNA (0.5)} \\
\cmidrule{2-13}
    & CD & EMD & CD & EMD & CD & EMD & CD & EMD & CD & EMD & CD & EMD \\
\midrule
    % NFD & 0.023 & 0.230 & 0.396 & 0.312 & 0.775 & 0.839 & 0.023 & 0.230 & 0.396 & 0.312 & {\bf 0.775} & 0.839 \\
    NFD~\cite{ndf2023shue} & 0.023 & 0.230 & 0.396 & 0.312 & 0.775 & 0.839 & - & - & - & - & - & - \\
    % PVD & 0.021 & 0.221 & 0.367 & 0.220 & 0.778 & 0.867 & {\bf 0.021} & 0.221 & 0.363 & 0.320 & 0.778 & 0.860 \\
    PVD~\cite{Zhou_2021_ICCV} & 0.021 & 0.221 & 0.367 & 0.220 & 0.778 & 0.867 & - & - & - & - & - & - \\
    SDFusion~\cite{cheng2023sdfusion} & 0.034 & 0.246 & 0.291 & 0.269 & 0.847 & 0.890 & 0.031 & 0.245 & 0.282 & 0.283 & 0.882 & 0.897 \\
    BlockFusion*~\cite{Wu2024blockfusion} & 0.048 & 0.305 & 0.177 & 0.110 & 0.953 & 0.986 & 0.054 & 0.330 & 0.186 & 0.091 & 0.961 & 0.993 \\
\midrule
    Ours & {\bf 0.019} & {\bf 0.140} & {\bf 0.421} & {\bf 0.316} & {\bf 0.738} & {\bf 0.512} & {\bf 0.021} & {\bf 0.147} & {\bf 0.410} & {\bf 0.321} & {\bf 0.783} & {\bf 0.458} \\
\bottomrule
\end{tabular}
}
\vspace{-0.2cm}
\caption{Geometric quality of synthesized 3D scene geometry as independent chunks (left) and as chunks of outpainted 3D scenes (right). \OURS{} generates scenes more reflective of ground-truth geometric distributions. 
*Note that BlockFusion results are generated unconditionally.
%Our generated scenes are more diverse and the proximity-based classifier is significantly confused between our and ground-truth scenes.
}
\label{tab:ind-eval}
\end{table}

\subsection{Evaluation Metrics}

We assess both generation and editing quality, in terms of geometric fidelity and adherence to text and editing inputs, evaluated for both individually generated chunks and crops of outpainted 3D scenes.
%on separate chunks conditioned on the input text queries and chunks which are generated to form a large-scale scene. 
%To evaluate editing results we perform user study assessment.

\textbf{Geometric quality.} We evaluate synthesized 3D scene geometry, following established evaluation metrics~\cite{pointflow2019yang, Wu2024blockfusion}, which do not take input conditions into account. 
Specifically, we use Minimum Matching Distance (MMD), Coverage (COV), and 1-Nearest-Neighbor-Accuracy (1-NNA). For MMD, lower is better; for COV, higher is better; for 1-NNA, 50\% is the optimal.
We use a Chamfer Distance (CD) distance measure for computing these metrics in 3D. Further details can be found in the supplementary.

\textbf{Consistency of synthesized geometry with text inputs.} To evaluate how well synthesized geometry corresponds to input text queries, we follow the evaluation proposed by ShapeGlot~\cite{achlioptas2019shapeglot}. A neural evaluator is trained to distinguish the target and distracting chunks, given the text description. 
% Given two chunks from different methods or one chunk from a method and another chunk from a GT set, the neural evaluator provides a confidence score for each of them based on the binary classification logits. % Comment this for arxiv/camera-ready
% If the absolute difference between two confidence scores $\leq 0.2$, we consider the comparison to be confused. % Comment this for arxiv/camera-ready

We also evaluate using CLIP~\cite{pmlr-v139-radford21a} score, which reflects the consistency of generated geometry to text inputs in CLIP space. We render each chunk from 5 views (1 top,  4 side views).
%, and compute the CLIP scores of the views to the chunk text caption. We choose a maximum score from the views to assign it to a chunk. 
Since individual views may contain occluded objects, we evaluate the max CLIP score. % to reflect text consistency.
%Among 5 views that we render for chunks, objects of some views can be blocked with walls, and thus the correlation of such a view to a text caption can be significantly lowered even if text-correlated objects are present on a scene chunk and visible in another view; therefore we use max CLIP score here.

\begin{table}[tp]
\centering
\resizebox{0.48\textwidth}{!}{
\begin{tabular}{cc|ccc|c}
\toprule
    Target (Tr) & Distractor (Dis) & P(Tr) & P(Dis) & P(conf.) & P(Dis=GT) - P(Tr) $\downarrow$ \\
\midrule
    Ours & SDFusion~\cite{cheng2023sdfusion} & 58\% & 42\% & 25\% & - \\
    Ours & Text2Room~\cite{hoellein2023text2room} & 65\% & 35\% & 32\% & - \\
\midrule
    SDFusion~\cite{cheng2023sdfusion} & GT & 33\% & 67\% & 25\% & 34\% \\
    Text2Room~\cite{hoellein2023text2room} & GT & 38\% & 62\% & 31\% & 24\% \\
    Ours & GT & 42\% & 58\% & 33\% & {\bf 16\%} \\
\bottomrule
\end{tabular}
}
\vspace{-0.25cm}
\caption{Quality of text-guided generation using a pretrained neural listener model.
Our results are preferred over that of SDFusion~\cite{cheng2023sdfusion}, and Text2Room~\cite{hoellein2023text2room}, both in direct comparison as well as relative to ground truth.
}
\label{tab:text-eval}
\vspace{-0.35cm}
\end{table}

\begin{table}[tp]
\centering
\resizebox{0.35\textwidth}{!}{
\begin{tabular}{l|c|c}
\toprule
    \multirow{1}{*}{Method} & \multicolumn{1}{c|}{Independent chunks} & \multicolumn{1}{c}{Scene chunks} \\
\midrule
    NFD~\cite{ndf2023shue} & 26.59 & 26.59 \\
    PVD~\cite{Zhou_2021_ICCV} & 24.79 & 24.79 \\
    SDFusion~\cite{cheng2023sdfusion} & 28.01 & 27.70 \\
\midrule
    Ours & {\bf 29.81} & {\bf 29.40} \\
\bottomrule
\end{tabular}
}
\vspace{-0.25cm}
\caption{CLIP-Score evaluation of text-guided generation. Rendered views of chunks generated by our method better match text captions.
}
\label{tab:clip-eval}
\vspace{-0.45cm}
\end{table}

% \begin{table}[tp]
% \centering
% \resizebox{0.4\textwidth}{!}{
% \begin{tabular}{l|c|c}
% \toprule
%     \multirow{1}{*}{Method} & \multicolumn{1}{c|}{Independent chunks} & \multicolumn{1}{c}{Scene chunks} \\
% \midrule
%     Text2Room~\cite{hoellein2023text2room} & 24.11 & 24.11 \\
%     Ours & {\bf 29.81} & {\bf 29.40} \\
% \bottomrule
% \end{tabular}
% }
% \caption{CLIP-Score evaluation of text-guided generation. Rendered views of chunks generated by our method better match text captions.
% }
% \label{tab:clip-eval-t2r}
% \vspace{-0.25cm}
% \end{table}

We also include a perceptual study in the supplemental.

% % User study
% \begin{figure}
% \begin{center}
%     \includegraphics[width=0.99\linewidth]{images/user_study.jpg}
%     \caption{User study for perceptual quality of text-guided 3D indoor scene generation and editing. (a) Unary study on perceptual geometric quality and text consistency for generated chunks and scenes. (b) Unary study on editing quality and scene consistency for \OURS{}. (c) Binary study between \OURS{} and baselines on text consistency between captions and generated chunks. (d) Binary study between \OURS{} and baselines on perceptual geometric quality of generated chunks. (e) Unary study of \OURS{} for locality of edits.
%     }
%     \label{fig:user_study}
% \end{center}
% \end{figure}

\subsection{Comparison with State of the Art}

We compare with several state-of-the-art 3D diffusion-based generative methods leveraging various geometry representations: PVD~\cite{Zhou_2021_ICCV} generates points, NFD~\cite{ndf2023shue} learns a latent triplane diffusion model, SDFusion~\cite{cheng2023sdfusion} leverages a scalable latent grid representation for text-conditioned generation, Text2Room~\cite{hoellein2023text2room} employs RGB image synthesis to fuse observations into a scene mesh, and BlockFusion~\cite{Wu2024blockfusion} uses a latent triplane diffusion model to generate large-scale scenes. We extend PVD and NFD approaches using the same BERT-based text encoding as SDFusion and ours.
%Since PVD and NFD were originally developed for unconditional generation, we use the same BERT-based text encoder as for our method for text-guided generation for PVD and NFD. 
We apply our scene outpainting strategy for PVD, NFD, and SDFusion, but find empirically that it fails to generate coherent scenes for PVD and NFD, so we visualize SDFusion; BlockFusion is designed to produce large-scale scenes using triplane outpainting. 

Tab.~\ref{tab:ind-eval} shows a quantitative evaluation of the geometric quality of generated chunks as well as chunks sampled from generated scenes for models trained with synthetically generated captions. 
Our factored approach produces consistently improved geometry in comparison with baselines. 
PVD~\cite{Zhou_2021_ICCV} does not decouple geometric compression and diffusion training and, using a limited number of points, which makes it unable to produce fine geometric details of scene chunks. NFD~\cite{ndf2023shue} struggles with the complex, diverse, non-canonicalized scene data.  In addition,  SDFusion~\cite{cheng2023sdfusion} and BlockFusion~\cite{Wu2024blockfusion} perform worse due to the lack of an intermediary spatially-structured condition.

This can also be seen in the qualitative results in Fig.~\ref{fig:scenes_qual}. BlockFusion uses latent triplanes to outpaint scenes; however, the triplanes can produce misshapen objects, especially those intersecting with outpainting seams. %To compare with BlockFusion, we generate chunks and scenes unconditionally since BlockFusion retrained with text conditioning performs highly unstable.
Note that following the authors' suggestions for comparisons with BlockFusion, the results are generated unconditionally without text input, in contrast to SDFusion and our method.
 
Tabs.~\ref{tab:text-eval} and \ref{tab:clip-eval} evaluate the consistency of generated geometry to the input text, showing that our factored approach better adheres to input text prompts.

In contrast to state-of-the-art 3D generative methods, \OURS{} enables localized editing of generated 3D scenes, as shown in Fig.~\ref{fig:editing}, maintaining scene consistency while manipulating geometry-based on easy box manipulations in the semantic domain. We include further comparisons and visualizations in the supplemental.

\begin{table}[tp]
\centering
\resizebox{0.48\textwidth}{!}{
\begin{tabular}{l|cc|cc|cc|cc|cc|cc}
\toprule
    \multirow{2}{*}{Method} & \multicolumn{6}{c|}{Independent chunks} & \multicolumn{6}{c}{Scene chunks} \\
\cmidrule{2-13}
    & \multicolumn{2}{c}{MMD $\downarrow$} & \multicolumn{2}{c}{COV $\uparrow$} & \multicolumn{2}{c|}{1-NNA (0.5)} & \multicolumn{2}{c}{MMD $\downarrow$} & \multicolumn{2}{c}{COV $\uparrow$} & \multicolumn{2}{c}{1-NNA (0.5)} \\
\cmidrule{2-13}
    & CD & EMD & CD & EMD & CD & EMD & CD & EMD & CD & EMD & CD & EMD \\
\midrule
    w/o sem stage & 0.022 & 0.146 & 0.405 & 0.295 & 0.757 & 0.482 & 0.026 & 0.231 & 0.409 & 0.299 & 0.785 & 0.895 \\
    w/o conv attn & 0.020 & 0.161 & {\bf 0.433} & 0.312 & 0.752 & 0.514 & 0.024 & 0.233 & {\bf 0.410} & {\bf 0.347} & 0.869 & 0.896 \\
    w/o 3D latent & 0.029 & 0.248 & 0.321 & 0.276 & 0.931 & 0.951 & 0.025 & 0.237 & 0.383 & 0.335 & 0.915 & {\bf 0.496} \\
\midrule
    Ours & {\bf 0.019} & {\bf 0.140} & 0.421 & {\bf 0.316} & {\bf 0.738} & {\bf 0.512} & {\bf 0.021} & {\bf 0.147} & {\bf 0.410} & 0.321 & {\bf 0.783} & 0.458 \\
\bottomrule
\end{tabular}
}
\vspace{-0.2cm}
\caption{Ablations. Our semantic proxy representation, 3D attention conditioning, and use of 3D latent spaces for semantics and geometry significantly improve generated scene quality.
}
\label{tab:ablation}
\vspace{-0.65cm}
\end{table}

\subsection{Ablation Studies}

\noindent
\textbf{What is the impact of using a proxy semantic map for 3D scene generation?} This helps to disentangle 3D scene generation into coarse object arrangement and refined geometric synthesis. Tab.~\ref{tab:ablation} shows that performance improves with a proxy semantic generation, which helps to avoid generating floaters and incoherent object geometry or arrangements, since the semantic map provides guidance for object type and extent. 
%Not using a proxy semantic map leads to generations with inconsistent objects of significantly lower quality and non-natural object arrangements. 

\smallskip
\noindent
\textbf{What is the effect of convolutional attention for geometric diffusion?} Tab.~\ref{tab:ablation} shows that attention with convolutions to extract queries, keys and values instead of linear layers and to interpret the semantic map enables far more accurate geometric synthesis than MLP-based attention, which tends to generate oversmoothed results. Convolutions with nontrivial window sizes enable better handling of correlations between a latent grid and condition due to encoding neighborhood information.

\smallskip
\noindent
\textbf{What is the impact of 3D latent grids for diffusion?}
We show in Tab.~\ref{tab:ablation} that using a 3D latent space for semantic map generation and geometric synthesis significantly improves generation over a 1D latent space. In contrast to a 1D latent space, a 3D latent space maintains spatial correlations to the semantic structure of the scene, producing more effective output scene geometry.
%Using 1D latent vectors instead of 3D latent grids is similar to the functioning of 1D attention modules because 1D attention needs to transform 3D latent into 1D versions to process with linear layers. In addition, we have here a 1D latent space that is not able to compress geometry as smoothly and efficiently as when using 3D latent. Thus, this leads to even more inconsistent and blobby object geometry compared to 3D attention ablation. The model can not properly attend to text information about object arrangements and scene layout, resulting in the effect of randomly fused geometric chunks when generating a scene.

\smallskip
\noindent \emph{Limitations.} \OURS{} offers a first step towards text-guided controllable 3D indoor scene generation, though various limitations remain. 
For instance, while input text to our method is flexibly encoded, we train our 3D semantic layouts on closed vocabulary data, which can limit generation diversity of generated object types.
Additionally, since we train on chunks and generate scenes by outpainting, room boundaries can be more difficult to control based on text input, instead requiring editing of the semantic map.
%Finally, generated scenes capture furniture-level object geometry, but resolution is limited by the input resolution of the VQ-VAE training.
%TODO
%For instance, our approach does not take additional text features into the second stage, where geometry is generated from semantic maps, but instead encodes the text in the form of bounding box labels and their sizes and aspect ratios, potentially leading to losing some secondary details from textual descriptions. Additionally, the method does not use single captions that describe scenes as a whole, demanding slightly more effort from users to provide chunkwise text descriptions.

%% file: sections/5_conclusion.tex
\section{Conclusion}

We have introduced \OURS{}, a new factored latent diffusion approach for controllable, editable 3D scene generation.
By disentangling the complex 3D scene generation task into first creating a coarse, high-level structural semantic, followed by finer-grained geometric refinement, \OURS{} enables both effective text-guided 3D scene synthesis of large-scale scenes, and moreover, synthesis of editable 3D scene representations.
Our coarse semantic map is structured as semantic boxes, enabling user-friendly box manipulation that can be used for various localized editing (object addition, removal, replacement, moving, altering size) of the generated 3D scenes.
We believe this represents an important step towards artist-driven automated 3D content creation, through the formulation of editable 3D scene generation for content creation scenarios.

%% file: sections/X_suppl.tex
\clearpage
\setcounter{page}{1}
\maketitlesupplementary

In this supplemental material, we provide details of data processing and caption generation in Section~\ref{sec:supp_data}, show the additional qualitative and quantitative comparison to diffusion- and non-diffusion-based methods in Section~\ref{sec:supp_qual}, provide details of the evaluation metrics and the perceptual study in Section~\ref{sec:supp_base} and additional implementation details in Section~\ref{sec:supp_impl}.

% Chunks 1
\begin{figure*}[bp]
\begin{center}
    \includegraphics[width=1.0\textwidth]{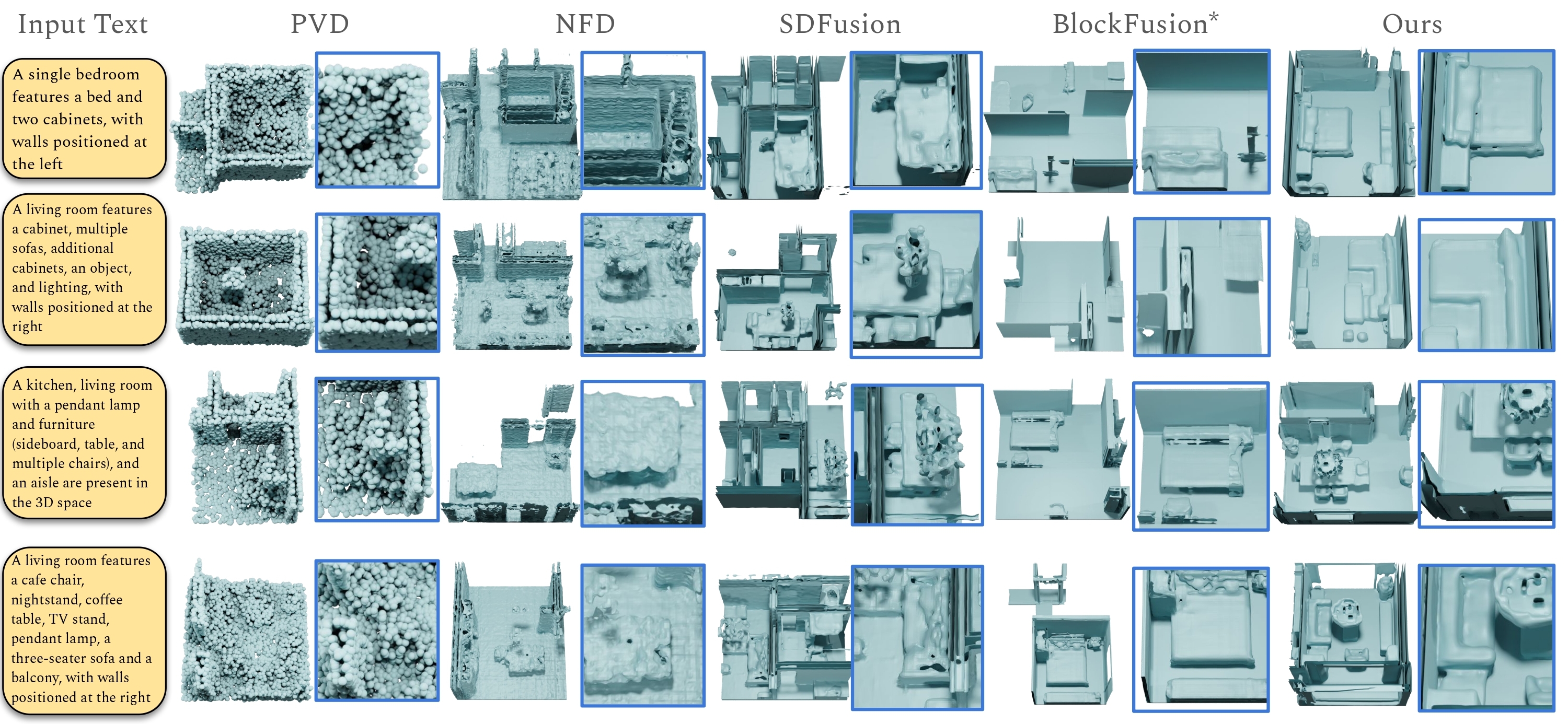}
    % \vspace{-0.4cm}
    \caption{
    Qualitative comparison with state of the art on text-guided scene chunk generation using Qwen1.5 captions. In comparison with PVD~\cite{Zhou_2021_ICCV}, NFD~\cite{ndf2023shue},  SDFusion~\cite{cheng2023sdfusion}, and BlockFusion~\cite{Wu2024blockfusion} \OURS{} generates higher-fidelity, more coherent scene structures through our factored approach.
    \\
    *Note that results for BlockFusion are generated unconditionally
    }
    \label{fig:chunks_supp_1}
\end{center}
\end{figure*}

\section{Data Processing}
\label{sec:supp_data}

\textbf{Geometry.} To make 3D-FRONT~\cite{fu20213dfront} data suitable for training and testing, we first combine 3D furniture and 3D scene meshes using 3D-FRONT annotation. 3D-FUTURE~\cite{fu20213dfuture} models are preliminarily converted into high-quality watertight meshes using the Manifold~\cite{huang2018robust} approach. This method can create meshes with double surfaces, so we remove all closed surfaces that lie within a mesh interior. To obtain the unsigned distance field of 3D-FRONT scenes with a resolution of 4.2 cm, we apply the virtual scanning tool mesh2sdf~\cite{Wang2022DualOG}. Preliminarily, we remove the ceiling from all 3D-FRONT scenes. In addition to the distance field, we regularly sample points with corresponding semantic labels belonging to scene layouts and furniture objects to form a semantic map of a scene with a resolution of 16.8 cm. The training chunks are obtained by randomly cropping from scene distance fields and semantic maps. We convert all test scenes into a test-suitable format by cutting the scenes into a regular grid of overlapping geometric and semantic chunks. All scene chunks are normalized to be centered at the origin and scaled to a unit cube.

% Editing
\begin{figure*}[tp]
\begin{center}
    \includegraphics[width=1.0\textwidth]{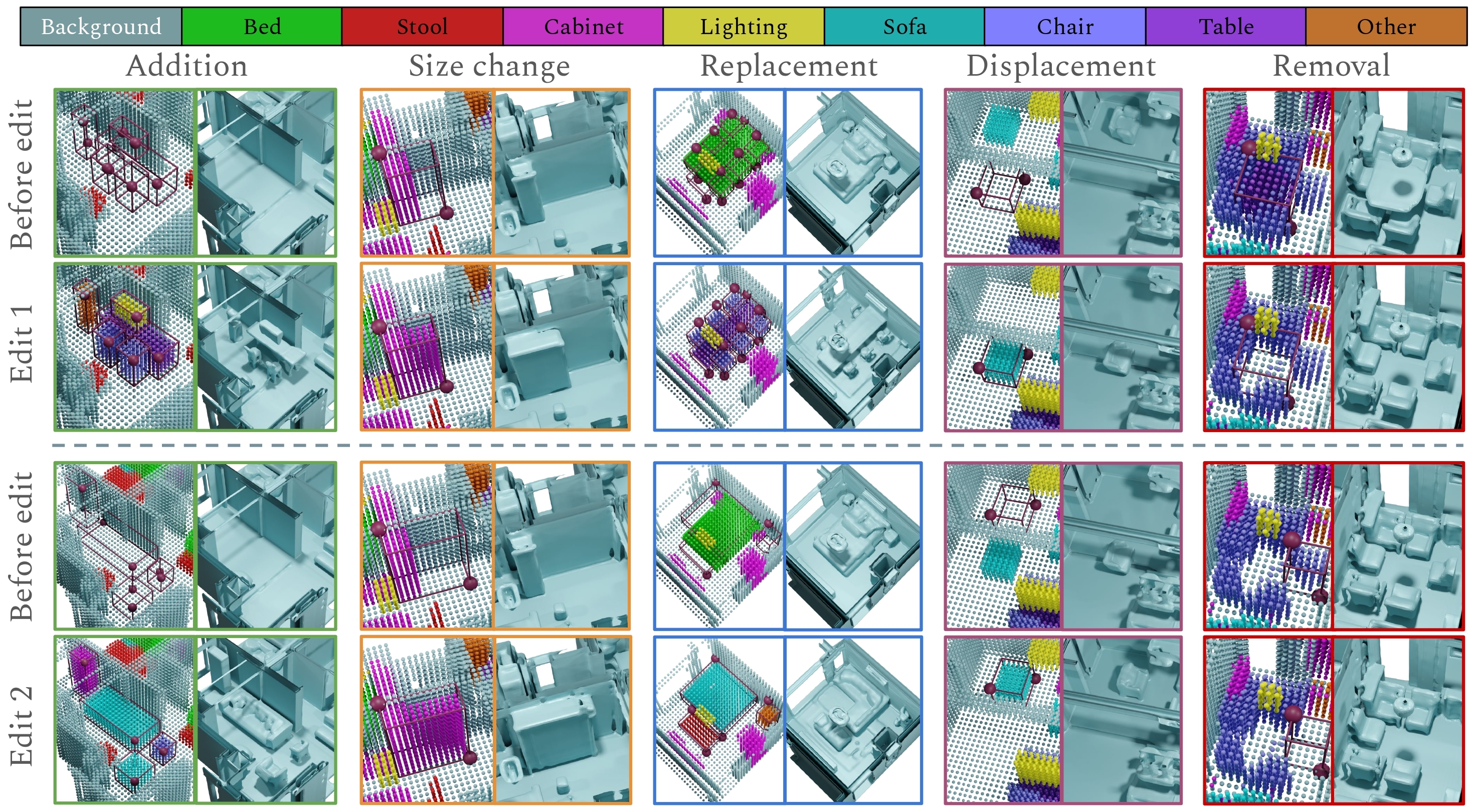}
    %\vspace{-0.7cm}
    \caption{Additional qualitative scene editing results. Generated scenes and their corresponding semantic maps are shown in the top row, and two alternatives for each object synthesis-based edit are shown below. 
    %\TODO{remove another object as well?}
    }
    \label{fig:edit_figonly}
\end{center}
\end{figure*}

\vspace{0.2cm}
\noindent \textbf{Captions.} To obtain captions for scene chunks, we use the 3D-FRONT object annotations to automatically generate seven types of captions. 
These caption types include descriptions with object counts or object lists without counts, subcategory information, and spatial relationships between objects.
First, every scene annotation includes object instances of 8 categories depicted in Fig.~\ref{fig:editing}. For every chunk, we add names of object categories into a caption if at least 35\% of an object lies within a chunk. Here, we have two types of text captions: explicit lists of single objects as category names and aggregated lists where repeated objects are counted. Another caption type can be obtained from the latter by adding spatial relationships between objects in a chunk. Second, using simple proximity checks based on Euclidean $L_2$ distance between object centers or object centers and wall points, we can identify if two or more objects form a group, stand across from each other, or stand next to a wall. For every caption, we also identify if there are walls along the borders of chunks. These three types of captions can be augmented using 33 subcategory names from 3D-FRONT annotation instead of category names. Finally, we have an extra room type caption, where for every chunk, we add room names from 3D-FRONT annotation to a caption if at least 25\% of a room lies in a chunk.

\emph{LLM-Refined Captions.}
Finally, we train additional instances of \OURS{}, SDFusion~\cite{cheng2023sdfusion}, NFD~\cite{ndf2023shue}, PVD~\cite{Zhou_2021_ICCV} with the second set of captions -- complex, natural text inputs. We utilize the large-language model Qwen1.5~\cite{qwen1.5} to refine our synthetic-looking captions using the following query: \texttt{Reformulate the following synthetic description of a 3D scene into a human-readable but concise, extremely minimalistic, and non-list format in only one sentence: <caption>}, where \texttt{<caption>} is the caption before LLM refinement.

\vspace{0.2cm}
\noindent \textbf{Augmentations.} During the training of the geometric and semantic VQ-VAE autoencoders and diffusion models, random $90^{\circ}$-fold rotation and symmetric reflection across $xz$- or $yz$-plane augmentations are applied to all train scene chunks and input latent representations.

% User study
\begin{figure*}[bp]
\begin{center}
    \includegraphics[width=0.92\textwidth]{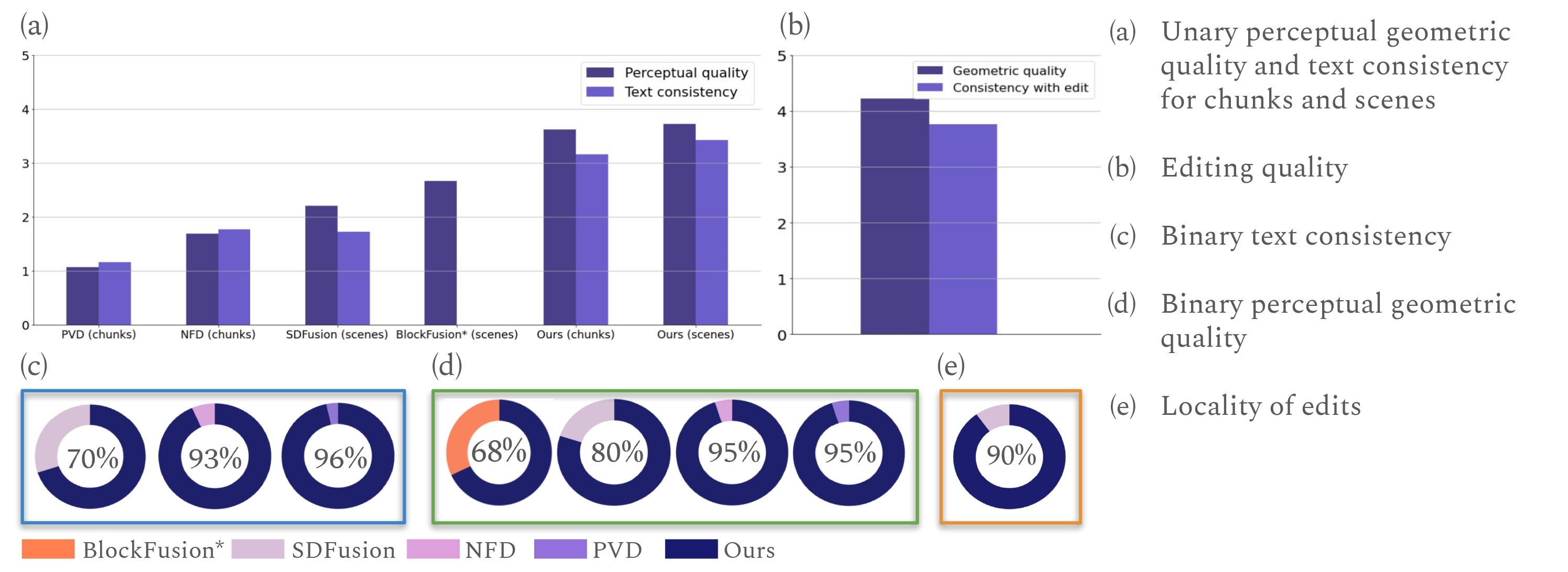}
    \caption{Perceptual study of the quality of text-guided 3D indoor scene generation and editing. (a) Unary study on perceptual geometric quality and text consistency for generated chunks and scenes. (b) Unary study on editing quality and scene consistency for \OURS{}. (c) Binary study between \OURS{} and baselines on text consistency between captions and generated chunks. (d) Binary study between \OURS{} and baselines on perceptual geometric quality of generated chunks. (e) Unary study of \OURS{} for locality of edits. 
    \\
    *Note that results for BlockFusion are generated unconditionally
    }
    \label{fig:user_study}
\end{center}
\end{figure*}

% \vspace{-18cm}
\section{Additional Results}
\label{sec:supp_qual}

\begin{figure*}
\begin{center}
    \includegraphics[width=0.85\textwidth]{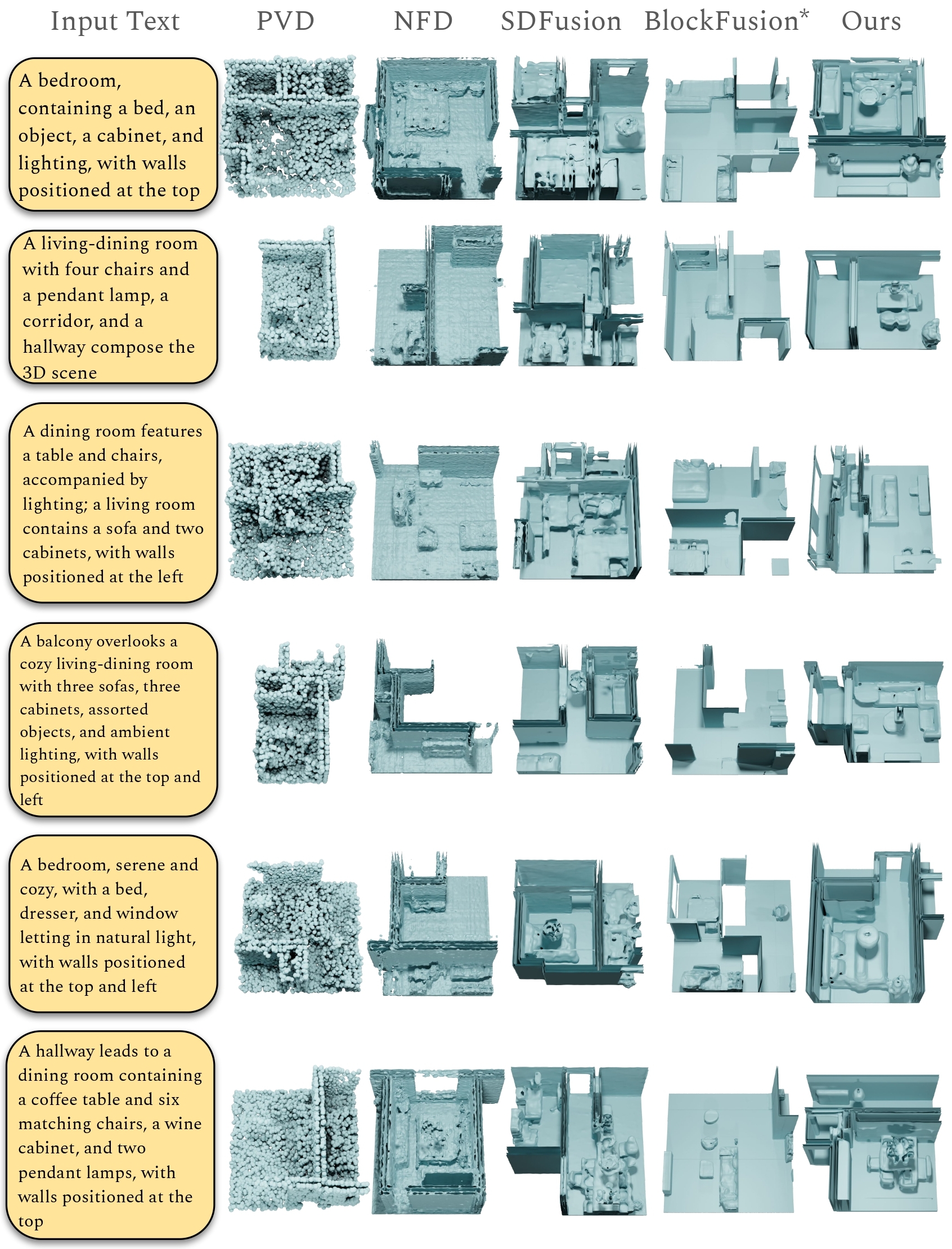}
    % \vspace{-0.4cm}
    \caption{
    Additional qualitative comparisons to state-of-the-art diffusion-based 3D generative approaches PVD~\cite{Zhou_2021_ICCV}, NFD~\cite{ndf2023shue}, SDFusion~\cite{cheng2023sdfusion}, and BlockFusion~\cite{Wu2024blockfusion} using Qwen1.5 captions. Our approach produces sharper scene geometry and more coherent scene structure.
    \\
    *Note that results for BlockFusion are generated unconditionally
    }
    \label{fig:chunks_supp_2}
\end{center}
\end{figure*}

\begin{figure*}
\begin{center}
    \includegraphics[width=0.85\textwidth]{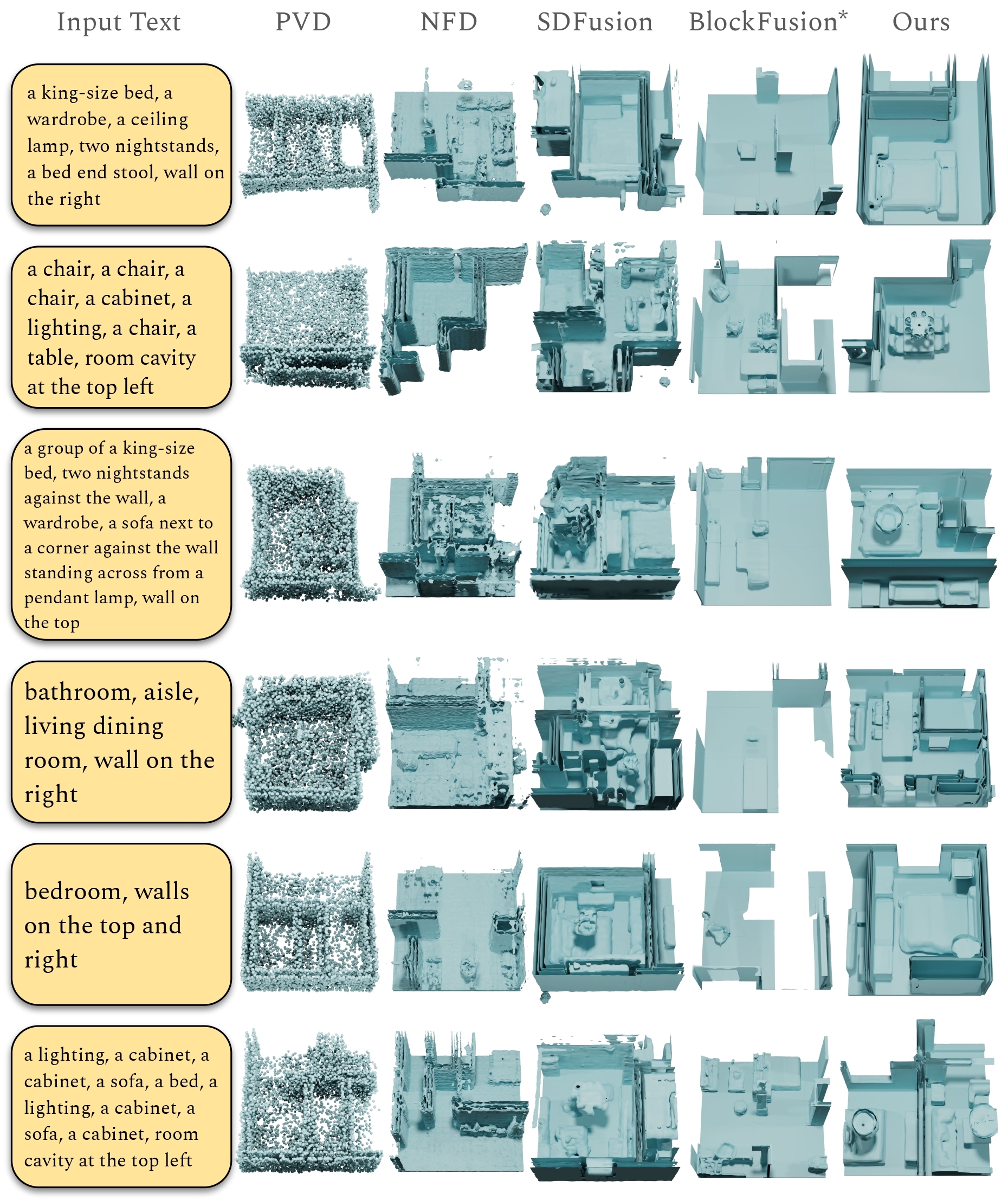}
    % \vspace{-0.4cm}
    \caption{
    Additional qualitative comparisons to state-of-the-art diffusion-based 3D generative approaches PVD~\cite{Zhou_2021_ICCV}, NFD~\cite{ndf2023shue}, SDFusion~\cite{cheng2023sdfusion}, and BlockFusion~\cite{Wu2024blockfusion} using synthetic captions. Our approach produces sharper scene geometry and more coherent scene structure.
    \\
    *Note that results for BlockFusion are generated unconditionally
    }
    \label{fig:chunks_supp_3}
\end{center}
\end{figure*}

\vspace{0.2cm}
\noindent \textbf{Additional Comparison to Diffusion-based Methods.} Fig.~\ref{fig:chunks_supp_1}, ~\ref{fig:chunks_supp_2} and~\ref{fig:chunks_supp_3} show additional qualitative comparisons with state-of-the-art baselines on scene chunk generation using synthetic and Qwen-refined captions. PVD~\cite{Zhou_2021_ICCV} model uses explicit point cloud diffusion, which makes it significantly harder to generate clean and complete scenes.
NFD~\cite{ndf2023shue} produces much cleaner scene layouts due to its signed distance field prediction. However, objects tend to lack details, with various low-level geometric artifacts due to the lack of structured latent space for generation. SDFusion~\cite{cheng2023sdfusion} can generate more recognizable furniture.
%that is recognizable as lamps, tables, chairs, beds, and cabinets. 
Nonetheless, due to direct text-to-geometry prediction and the absence of convolutional attention, SDFusion tends to generate more incoherent global structures (e.g., objects penetrating each other and inconsistent walls). Finally, BlockFusion~\cite{Wu2024blockfusion} unconditional generations contain inconsistent wall structures, and triplane-based generation is unable to produce accurate furniture objects in arbitrary chunk locations. In Tab.~\ref{tab:ind-eval-qwen-supp},~\ref{tab:clip-eval-qwen-supp}, we provide the quantitative evaluation of our method and baseline approaches for the geometric quality and text-guided generation using Qwen1.5 captions as input.

\begin{figure*}
\begin{center}
    \includegraphics[width=1.05\textwidth]{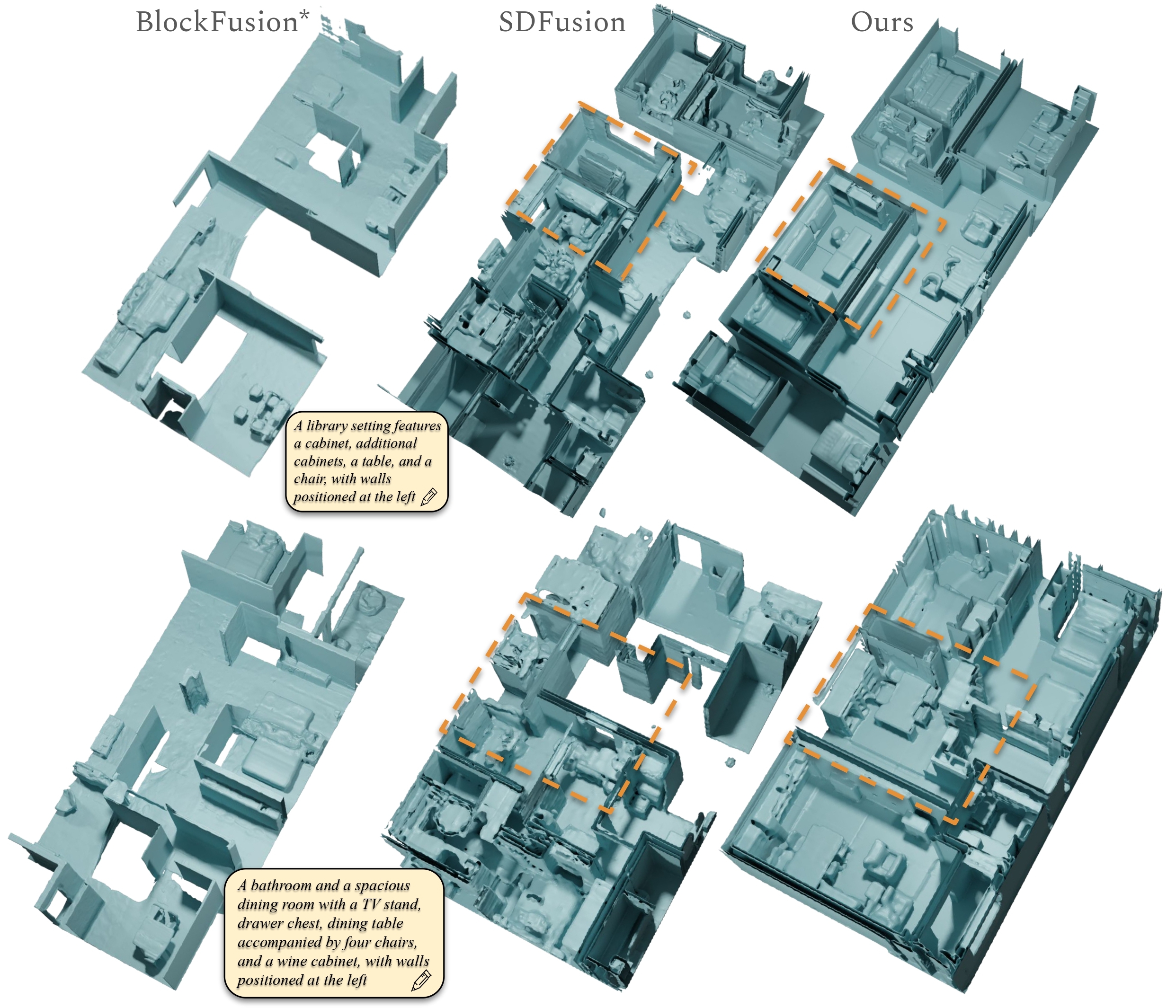}
    % \vspace{-0.7cm}
    \caption{
    Additional qualitative comparisons for scene generation in comparison with SDFusion~\cite{cheng2023sdfusion} and BlockFusion~\cite{Wu2024blockfusion}.
    \\
    *Note that results for BlockFusion are generated unconditionally
    }
    \label{fig:scene_supp_1}
\end{center}
\end{figure*}

\begin{figure*}
\begin{center}
    \includegraphics[width=1.05\textwidth]{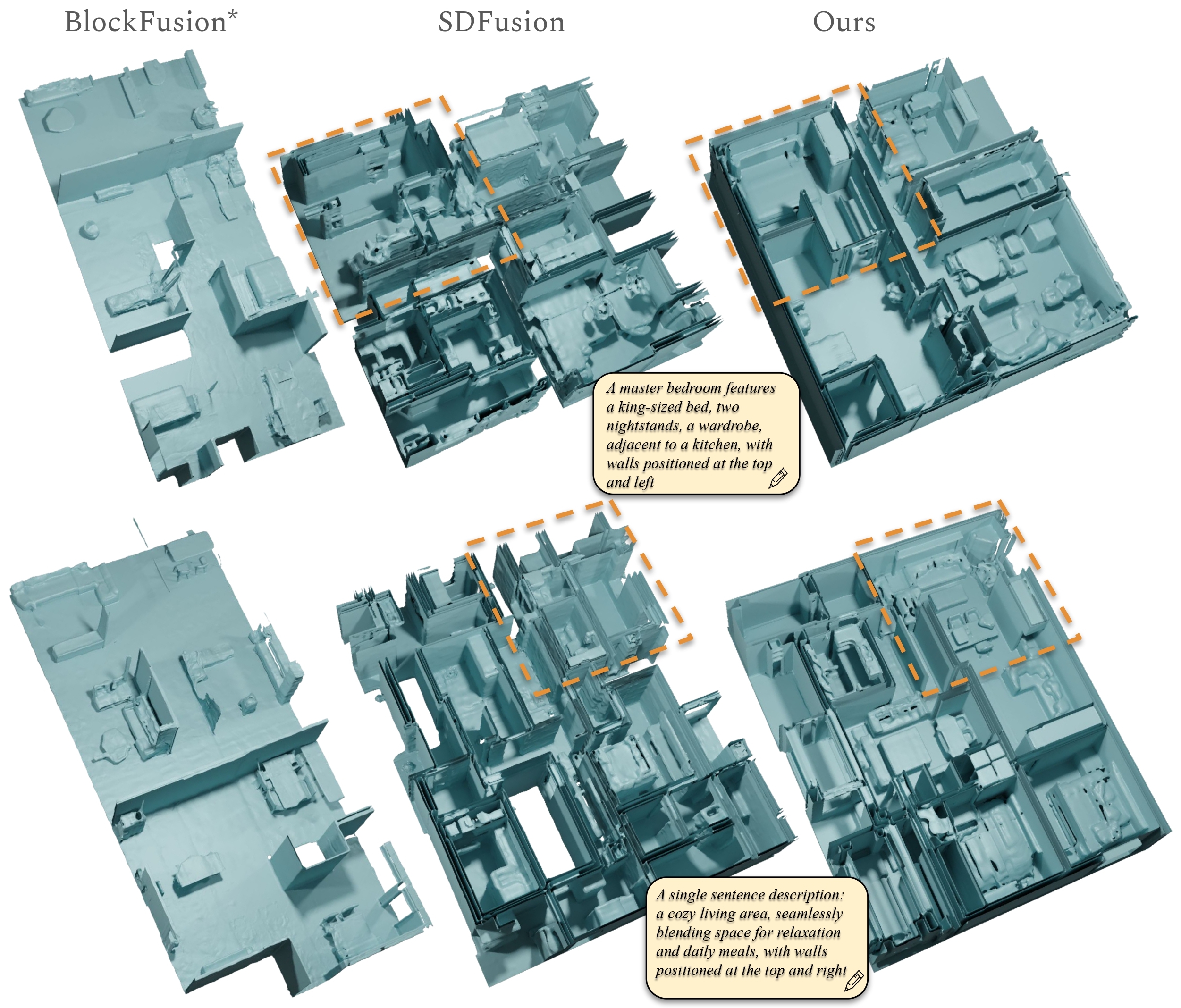}
    % \vspace{-0.7cm}
    \caption{
    Additional qualitative comparisons for scene generation in comparison with SDFusion~\cite{cheng2023sdfusion}  and BlockFusion~\cite{Wu2024blockfusion}. 
    \\
    *Note that results for BlockFusion are generated unconditionally
    }
    \label{fig:scene_supp_2}
\end{center}
\end{figure*}

Figs.~\ref{fig:scene_supp_1} and~\ref{fig:scene_supp_2} show additional qualitative comparisons for 3D scene generation with SDFusion~\cite{cheng2023sdfusion} and BlockFusion~\cite{Wu2024blockfusion}. 
SDFusion tends to produce more noticeable transitions between generated chunks, along with floating geometric artifacts and holes in furniture objects. Both SDFusion and BlockFusion generate significant artifacts, such as holes in the floor, due to the lack of conditioning on spatial information. BlockFusion struggles to outpaint objects from one chunk to the next chunk, which results in a significantly unnatural appearance of the generated room spaces.

Finally, we provide additional qualitative scene editing results for our method in Fig.~\ref{fig:edit_figonly}. Our approach is able to produce diverse and consistent editing results for the same input scene.

\vspace{0.2cm}
\noindent \textbf{Comparison to Non-diffusion-based Methods.} In addition, we provide a comparison to 2D diffusion lifting-based approach Text2Room~\cite{hoellein2023text2room} and a retrieval-based method ATISS~\cite{Paschalidou2021NEURIPS}, for which we evaluate only independent chunks generation since these models are not applicable for large-scale scene generation.

Tab.~\ref{tab:supp-ind-eval-t2r} quantitatively evaluates the geometric quality of generated chunks against ATISS and Text2Room. Text2Room~\cite{hoellein2023text2room} takes significant time to generate one chunk ($\sim$ 3.5 hours); therefore, we limited the evaluation of this approach to 92 chunks. Our factored approach produces consistently improved geometry in comparison with these baselines. In Tab.~\ref{tab:supp-clip-eval-t2r}, we also show that our approach significantly outperforms Text2Room by the CLIP score between rendered chunks and input text captions. We do not evaluate against the retrieval-based ATISS method because the CLIP score is biased towards non-generated but retrieved synthetic meshes placed on top of the floor. We found this comparison not meaningful. Instead, we evaluate our approach against ATISS using a pretrained neural listener model for the input text correspondence in Tab.~\ref{tab:text-eval-supp}. 
A neural evaluator is trained to distinguish the target chunk from a distracting chunk, given the text description. % Comment this for paper submission
Given two chunks from different methods or one chunk from a method and another chunk from a GT set, the neural evaluator provides a confidence score for each of them based on the binary classification logits. % Comment this for paper submission
If the absolute difference between two confidence scores $\leq 0.2$, we consider the comparison to be confused. % Comment this for paper submission
ATISS is not able to handle a large diversity of text captions and is significantly inferior to our approach in terms of text coherence.

\vspace{0.2cm}
\noindent \textbf{Additional Semantic Evaluation.} We provide additional analysis of our first-stage semantic map generation model, where the original latent diffusion model and the diffusion model explicitly trained with one-hot semantic maps are compared to each other.
We first compute the average chunk semantic accuracy with respect to text input, where for every object class category mentioned in a caption, we check if the corresponding object has been predicted. For this metric, the latent-based model has accuracy of ${\bf 91\%}$ against $83\%$ for the model without latent representation. In Tab.~\ref{tab:supp-sem}, we provide further evaluation, which is based on MMD/COV/1-NNA metrics.

\section{Baseline Evaluation Setup}
\label{sec:supp_base}

\noindent \textbf{Metrics.} Following the works for 3D shape generation, we use the following metrics on point clouds extracted from mesh surfaces:

$$\text{MMD}(S_g, S_r) = \frac{1}{|S_r|}\sum_{Y\in S_r}\min_{X\in S_g}D(X,Y),$$

$$\text{COV}(S_g, S_r) = \frac{|\{\text{argmin}_{Y\in S_r} D(X,Y)|X\in S_g\}|}{|S_r|},$$

$$\text{1-NNA}(S_g, S_r) = \frac{\sum_{X\in S_g} \mathbf{1}_X + \sum_{Y\in S_r} \mathbf{1}_Y}{|S_g| + |S_r|},$$

$$\mathbf{1}_X = \mathbf{1}[N_X \in S_g],\ \mathbf{1}_Y = \mathbf{1}[N_Y \in S_r],$$
where $S_r$ and $S_g$ are reference and generated sets of point clouds extracted from ground-truth and generated mesh surfaces, respectively, $N_X$ is a point cloud that is closest to $X$ in both generated and reference dataset, i.e., $N_X = \text{argmin}_{K\in S_r \cup S_g} D(X,K)$. We use Chamfer distance (CD) and Earth-mover distance (EMD) as $D(X, Y)$ to compute these metrics in 3D. To evaluate these metrics, we extract 4096 points from ground-truth and generated mesh surfaces or sample 4096 points from PVD point clouds.

We utilize the official implementations of NFD~\cite{ndf2023shue}, PVD~\cite{Zhou_2021_ICCV}, SDFusion~\cite{cheng2023sdfusion}, BlockFusion~\cite{Wu2024blockfusion}, Text2Room~\cite{hoellein2023text2room}, and ATISS~\cite{Paschalidou2021NEURIPS}. For NFD and PVD, we do not implement the same or similar scene-aware generation mechanism, which inpaints missing chunks because PVD leverages the explicit point cloud representation in the diffusion model, and NFD demonstrates extremely poor results when using an inpaiting mechanism resulting in empty chunks which degrade in quality along the generation sequence. Text2Room and ATISS approaches are also inapplicable for large-scale scene generation using the outpainting mechanism. We use the same context encoding for text captions as in \OURS~ and SDFusion for NFD and PVD, while Text2Room, ATISS, and BlockFusion are designed to take text as input.

To evaluate geometric quality in Tab.~\ref{tab:ind-eval}, we normalize the ground-truth and predicted chunk meshes or point clouds into a unit cube and extract 4096 points from mesh surface or point cloud. 

For the text-aware evaluation in Tab.~\ref{tab:text-eval}, we train the neural listener model consisting of geometric encoder, text embedded, and language encoder. Geometric encoder consists of 5 ResNet blocks with GeLU activations and 2 linear layers with ReLU activations and takes uDF of geometric chunks as input. The input text is encoded using the same text encoder as in \OURS, but with an embedding dimension of 128. The text features are then processed using the LSTM~\cite{hochreiter1997lstm} network. The resulting features are concatenated with geometric features and finally processed with a shallow MLP network with ReLU activations.

For the CLIP score evaluation in Tab.~\ref{tab:clip-eval}, we render 4 views of predicted meshes or point clouds and compute the cosine distance to text caption used for generation. We add a prefix 'a render of a 3D scene with ' to captions for CLIP score evaluation for ones not generated with Qwen1.5~\cite{qwen1.5} model.

\vspace{0.2cm}
\noindent \textbf{Perceptual Study.}
To more effectively capture the perceptual quality of synthesized geometry, as well as adherence to text and editing inputs, we perform a perceptual study. We ask users to evaluate perceptual geometric quality as well as adherence to the text prompts, both as unary evaluation scores and binary comparisons between \OURS{} and each baseline.
Perceptual geometric quality is assessed on a scale from 1 (Awful quality) to 5 (Great quality). Adherence to text input is assessed on a scale from 1 (Not matching) to 5 (Matching).

In particular, since we lack ground truth editing results as well as baselines that perform local spatial edits, we evaluate our editing performance through unary evaluation in the perceptual study.
Editing results in the perceptual study are generated randomly across each possible editing operation. We ask users to assess (1) if the resulting edited scene is consistent with the given edit operation using a scale from 1 to 5; (2) the perceptual geometric quality of an edited scene using a scale from 1 to 5; and (3) if a scene remained unchanged outside of the editing region as either 1 (Yes) or 2 (No). In total, 21 participants took part in a perceptual study consisting of 53 questions per user. We provide the quantitative results of the conducted perceptual study in Fig.~\ref{fig:user_study}.

We developed a Django-based web application for the perceptual study. In total, we have 5 sections for our survey. For the first part, an unary study on perceptual geometric quality and text consistency for generated chunks and scenes, there are 25 questions and 5 randomly chosen scenes and chunks for every approach. Here, the user is asked to provide a score from 1 to 5 based on the perceptual geometric quality of chunks and the consistency of generation to an input text caption. In addition to chunkwise comparison, for SDFusion, BlockFusion, and our approach, there is also a unary study on scenes, where users are asked to evaluate the geometric quality of the whole scene. For SDFusion and ours, users are asked to evaluate the consistency of one scene chunk to a text caption.

\section{Implementation Details}
\label{sec:supp_impl}
Our method is implemented using PyTorch. Semantic and geometric VQ-VAE models are trained with an Adam~\cite{KingBa15} optimizer with learning rates 1e-4 and 2e-4 for the semantic and geometric VQ-VAEs. We use AdamW~\cite{Loshchilov2017DecoupledWD} with a learning rate of 1e-5 for both semantic and geometric latent diffusion models. The semantic and geometric VQ-VAEs are trained on 2 NVIDIA A6000s each for 320k and 160k iterations ($\sim$ 50 hours) until convergence. The diffusion models are trained on 2 NVIDIA A100s each for 400k iterations ($\sim$ 100 and 150 hours, respectively).

VQ-VAE semantic and geometric models comprise 3 ResNet blocks in the encoder and 3 ResNet blocks in the decoder with bilinear upsampling layers and GeLU~\cite{hendrycks2016gelu} nonlinearities. For the semantic VQ-VAE latent space, we encode semantic chunks into $(1, 4, 4, 4)$ latent grids, with only 1 feature channel using a dictionary size of 8192. Geometric chunks are encoded using the geometric VQ-VAE model into $(1, 16, 16, 16)$ latent grids, with 1 feature channel using a dictionary size of 32768. 

The semantic diffusion model is trained using larger latent grids of size $(1, 8, 4, 8)$ that correspond to two twice bigger semantic chunks in both horizontal dimensions. We pad these grids with zeros to the shape of $(1, 8, 8, 8)$ to enable compression using 4 ResNet blocks in the encoder of the UNet model. The first 3 ResNet blocks combine convolutional operations with attention layers with 8 heads. To encode the context, we use the transformer-based model with BERT tokenizer and context dimension of 1280 and 77 maximum number of tokens.

Analogously, the geometric diffusion model is trained using larger latent grids of size $(1, 32, 16, 32)$ that correspond to two twice bigger geometric chunks in both horizontal dimensions. The UNet model encoder consists of 3 ResNet blocks with attention layers with 8 heads in each block. To encode the semantic context, we first encode the input semantic chunk of size $(1, 32, 16, 32)$ into one-hot representation with 10 class channels. This one-hot representation is encoded into a context feature grid of size $(128, 16, 8, 16)$ using the fully convolutional network with LeakyReLU activations~\cite{bing2015lrelu}.

\begin{table}[tp]
\centering
\resizebox{0.48\textwidth}{!}{
\begin{tabular}{l|cc|cc|cc}
\toprule
    \multirow{2}{*}{Method} & \multicolumn{6}{c}{Independent chunks} \\
\cmidrule{2-7}
    & \multicolumn{2}{c}{MMD $\downarrow$} & \multicolumn{2}{c}{COV $\uparrow$} & \multicolumn{2}{c}{1-NNA (0.5)} \\
\cmidrule{2-7}
    & CD & EMD & CD & EMD & CD & EMD \\
\midrule
    w/o latent & 0.263 & 0.473 & 0.335 & 0.344 & 0.784 & 0.784 \\
\midrule
    Ours & {\bf 0.222} & {\bf 0.458} & {\bf 0.495} & {\bf 0.491} & {\bf 0.598} & {\bf 0.631} \\
\bottomrule
\end{tabular}
}
\caption{Semantic quality of synthesized 3D scene geometry as independent chunks.
}
\label{tab:supp-sem}
\end{table}

\begin{table}[bp]
\centering
\resizebox{0.48\textwidth}{!}{
\begin{tabular}{l|cc|cc|cc}
\toprule
    \multirow{2}{*}{Method} & \multicolumn{6}{c}{Independent chunks} \\
\cmidrule{2-7}
    & \multicolumn{2}{c}{MMD $\downarrow$} & \multicolumn{2}{c}{COV $\uparrow$} & \multicolumn{2}{c}{1-NNA (0.5)} \\
\cmidrule{2-7}
    & CD & EMD & CD & EMD & CD & EMD \\
\midrule
    Text2Room & 0.048 & 0.316 & 0.021 & 0.021 & 0.997 & 0.993 \\
    ATISS~\cite{Paschalidou2021NEURIPS} & 0.050 & 0.327 & 0.117 & 0.117 & 0.993 & 0.992 \\
\midrule
    Ours & {\bf 0.019} & {\bf 0.140} & {\bf 0.421} & {\bf 0.316} & {\bf 0.738} & {\bf 0.512} \\
\bottomrule
\end{tabular}
}
\caption{Geometric quality of synthesized 3D scene geometry as independent chunks (left) and as chunks of outpainted 3D scenes (right).
}
\label{tab:supp-ind-eval-t2r}
\end{table}

\begin{table}[tp]
\centering
\resizebox{0.48\textwidth}{!}{
\begin{tabular}{l|c|c}
\toprule
    \multirow{1}{*}{Method} & \multicolumn{1}{c|}{Independent chunks} & \multicolumn{1}{c}{Scene chunks} \\
\midrule
    Text2Room~\cite{hoellein2023text2room} & 24.11 & 24.11 \\
    Ours & {\bf 29.81} & {\bf 29.40} \\
\bottomrule
\end{tabular}
}
\caption{CLIP-Score evaluation of text-guided generation. Rendered views of chunks generated by our method better match text captions.
}
\label{tab:supp-clip-eval-t2r}
\vspace{-0.25cm}
\end{table}

\begin{table}[tp]
\centering
\resizebox{0.48\textwidth}{!}{
\begin{tabular}{cc|ccc|c}
\toprule
    Target (Tr) & Distractor (Dis) & P(Tr) & P(Dis) & P(conf.) & P(Dis=GT) - P(Tr) $\downarrow$ \\
\midrule
    Ours & ATISS~\cite{Paschalidou2021NEURIPS} & 66\% & 34\% & 21\% & - \\
\midrule
    ATISS~\cite{Paschalidou2021NEURIPS} & GT & 33\% & 67\% & 22\% & 34\% \\
    Ours & GT & 42\% & 58\% & 33\% & {\bf 16\%} \\
\bottomrule
\end{tabular}
}
\vspace{-0.25cm}
\caption{Quality of text-guided generation using a pretrained neural listener model.
Our results are preferred over that of SDFusion~\cite{cheng2023sdfusion}, ATISS~\cite{Paschalidou2021NEURIPS}, and Text2Room~\cite{hoellein2023text2room}, both in direct comparison as well as relative to ground truth.
}
\label{tab:text-eval-supp}
\vspace{-0.25cm}
\end{table}

\begin{table}[bp]
\centering
\resizebox{0.48\textwidth}{!}{
\begin{tabular}{l|cc|cc|cc|cc|cc|cc}
\toprule
    \multirow{2}{*}{Method} & \multicolumn{6}{c|}{Independent chunks} & \multicolumn{6}{c}{Scene chunks} \\
\cmidrule{2-13}
    & \multicolumn{2}{c}{MMD $\downarrow$} & \multicolumn{2}{c}{COV $\uparrow$} & \multicolumn{2}{c|}{1-NNA (0.5)} & \multicolumn{2}{c}{MMD $\downarrow$} & \multicolumn{2}{c}{COV $\uparrow$} & \multicolumn{2}{c}{1-NNA (0.5)} \\
\cmidrule{2-13}
    & CD & EMD & CD & EMD & CD & EMD & CD & EMD & CD & EMD & CD & EMD \\
\midrule
    NFD~\cite{ndf2023shue} & 0.023 & 0.225 & {\bf 0.411} & {\bf 0.335} & 0.744 & 0.814 & - & - & - & - & - & - \\
    PVD~\cite{Zhou_2021_ICCV} & {\bf 0.021} & 0.221 & 0.396 & 0.285 & {\bf 0.729} & 0.876 & - & - & - & - & - & - \\
    SDFusion~\cite{cheng2023sdfusion} & 0.031 & 0.240 & 0.331 & 0.277 & 0.835 & 0.898 & 0.035 & 0.253 & 0.313 & 0.265 & 0.874 & {\bf 0.910} \\
    BlockFusion*~\cite{Wu2024blockfusion} & 0.048 & 0.305 & 0.177 & 0.110 & 0.953 & 0.986 & 0.054 & 0.330 & 0.186 & 0.091 & 0.961 & 0.993 \\
\midrule
    Ours & {\bf 0.021} & {\bf 0.165} & 0.399 & 0.300 & 0.772 & {\bf 0.769} & {\bf 0.026} & {\bf 0.249} & {\bf 0.365} & {\bf 0.270} & {\bf 0.839} & {\bf 0.910} \\
\bottomrule
\end{tabular}
}
\vspace{-0.2cm}
\caption{Geometric quality of synthesized 3D scene geometry as independent chunks (left) and as chunks of outpainted 3D scenes (right) generated with Qwen1.5 captions.
}
\label{tab:ind-eval-qwen-supp}
\end{table}

\begin{table}[tp]
\centering
\resizebox{0.48\textwidth}{!}{
\begin{tabular}{l|c|c}
\toprule
    \multirow{1}{*}{Method} & \multicolumn{1}{c|}{Independent chunks} & \multicolumn{1}{c}{Scene chunks} \\
\midrule
    NFD~\cite{ndf2023shue} & 21.61 & 21.61 \\
    PVD~\cite{Zhou_2021_ICCV} & 20.32 & 20.32 \\
    SDFusion~\cite{cheng2023sdfusion} & 23.08 & 23.17 \\
\midrule
    Ours & {\bf 23.96} & {\bf 23.79} \\ % 23.52 23.96 | 23.60 23.79
\bottomrule
\end{tabular}
}
\vspace{-0.25cm}
\caption{CLIP-Score evaluation of text-guided generation using Qwen1.5 captions. Rendered views of chunks generated by our method better match text captions.
}
\label{tab:clip-eval-qwen-supp}
\vspace{-0.25cm}
\end{table}

% Scene 1

% Scene 2

% Chunks 2

% Chunks 3
